\documentclass[sigconf,screen,preprint]{acmart} 

\usepackage[nolist]{acronym}
\usepackage{xspace}
\usepackage{todonotes} 
\usepackage{graphicx}
\usepackage{subcaption} 

\newcommand{\ie}{i.e.\xspace}
\newcommand{\eg}{e.g.\xspace}
\newcommand{\etal}{et al.\xspace}

\newcommand{\nDecisions}{15\xspace}

\newcommand{\opfgym}{\textit{OPF-Gym}\xspace}

\newcommand{\optuna}{\textit{Optuna}\xspace}

\newcommand{\voltageControl}{\textit{Voltage Control}\xspace}
\newcommand{\ecoDispatch}{\textit{Economic Dispatch}\xspace}
\newcommand{\loadShedding}{\textit{Load Shedding}\xspace}
\newcommand{\maxRenewable}{\textit{Max Renewables}\xspace}
\newcommand{\qmarket}{\textit{Q-Market}\xspace}

\newcommand{\validReward}{\textit{Valid Reward}\xspace}
\newcommand{\invalidPenalty}{\textit{Invalid Penalty}\xspace}
\newcommand{\objectiveShare}{\textit{Invalid Objective Share}\xspace}
\newcommand{\penaltyWeight}{\textit{Penalty Weight}\xspace}
\newcommand{\diffObjective}{\textit{Diff-Objective}\xspace}

\newcommand{\normalShare}{\textit{Normal Data}\xspace}
\newcommand{\uniformShare}{\textit{Uniform Data}\xspace}
\newcommand{\simbenchShare}{\textit{Realistic Data}\xspace}

\newcommand{\addVoltageMag}{\textit{Add Voltage Magnitude}\xspace}
\newcommand{\addVoltageAngle}{\textit{Add Voltage Angle}\xspace}
\newcommand{\addLineLoad}{\textit{Add Line Loading}\xspace}
\newcommand{\addTrafoLoad}{\textit{Add Trafo Loading}\xspace}
\newcommand{\addSlackPower}{\textit{Add Slack Power}\xspace}

\newcommand{\nSteps}{\textit{Steps Per Episode}\xspace}

\newcommand{\actionAutoscale}{\textit{Autoscaling}\xspace}

\newcommand{\meanError}{\textit{mean error}\xspace}
\newcommand{\invalidShare}{\textit{invalid share}\xspace}

\newcommand{\objective}{\textit{Optimization}\xspace}
\newcommand{\constraints}{\textit{Validity}\xspace}
\newcommand{\utopia}{\textit{Utopia}\xspace}
\newcommand{\pareto}{\textit{Pareto}\xspace}
\AtBeginDocument{%
  }

\setcopyright{acmlicensed}
\copyrightyear{2025}
\acmYear{2025}
\acmDOI{XXXXXXX.XXXXXXX}
\acmConference[]{ACM e-Energy 2025}{June 17--20,
  2025}{Rotterdam, Netherlands}
\acmISBN{978-1-4503-XXXX-X/18/06}

\begin{document}


\title{A General Approach of Automated Environment Design for Learning the Optimal Power Flow}

\author{Thomas Wolgast}
\email{thomas.wolgast@uni-oldenburg.de}
\orcid{0000-0002-9042-9964}
\affiliation{%
  \institution{Carl von Ossietzky Universität Oldenburg}
  \city{Oldenburg}
  \country{Germany}
}

\author{Astrid Nieße}
\email{astrid.niesse@uni-oldenburg.de}
\orcid{0000-0003-1881-9172}
\affiliation{%
  \institution{Carl von Ossietzky Universität Oldenburg}
  \city{Oldenburg}
  \country{Germany}
}

\renewcommand{\shortauthors}{Wolgast and Nieße}

\begin{abstract}
Reinforcement learning (RL) algorithms are increasingly used to solve the optimal power flow (OPF) problem. Yet, the question of how to design RL environments to maximize training performance remains unanswered, both for the OPF and the general case. We propose a general approach for automated RL environment design by utilizing multi-objective optimization. For that, we use the hyperparameter optimization (HPO) framework, which allows the reuse of existing HPO algorithms and methods. On five OPF benchmark problems, we demonstrate that our automated design approach consistently outperforms a manually created baseline environment design. Further, we use statistical analyses to determine which environment design decisions are especially important for performance, resulting in multiple novel insights on how RL-OPF environments should be designed. Finally, we discuss the risk of overfitting the environment to the utilized RL algorithm. To the best of our knowledge, this is the first general approach for automated RL environment design.\footnote{All code and data are open source and can be found at: https://github.com/Digitalized-Energy-Systems/auto-env-design.}
\end{abstract}

\received{20 February 2007}
\received[revised]{12 March 2009}
\received[accepted]{5 June 2009}

\maketitle
\begin{acronym}
    \acro{RL}[RL]{Reinforcement Learning}
    \acro{OPF}{Optimal Power Flow}
    \acro{HV}{High-Voltage}
    \acro{MV}{Medium-Voltage}
    \acro{LV}{Low-Voltage}
    \acro{DDPG}{Deep Deterministic Policy Gradient}
    \acro{SAC}{Soft Actor Critic}
    \acro{PPO}{Proximal Policy Optimization}
    \acro{TD3}{Twin-Delayed DDPG}
    \acro{ANN}{Artificial Neural Network}
    \acro{MAPE}{Mean Absolute Percentage Error}
    \acro{MDP}{Markov Decision Process}
    \acroplural{MDP}[MDPs]{Markov Decision Processes}
    \acro{HPO}{Hyperparameter Optimization}
    \acro{MINLP}{Mixed Integer Nonlinear Programming}
\end{acronym}

\section{Introduction}\label{sec:intro}
The general framework for the optimization of power grid states is the \ac{OPF} \cite{frankOptimalPowerFlow2012}, which aims to optimize power grid states subject to various constraints like voltage or line load constraints.  
However, conventional solvers can become very slow for more complex variants of the \ac{OPF} \cite{kotaryEndtoEndConstrainedOptimization2021}, which restricts use cases where \ac{OPF} solutions are required in high frequency for various different situations. For example, that is the case for real-time operation of power systems \cite{yanRealTimeOptimalPower2020} or in simulations where millions of different \ac{OPF} scenarios are investigated \cite{wolgastReinforcementLearningVulnerability2021}.
\par 
One emerging approach for faster \ac{OPF} calculation is to train deep neural networks to learn the mapping from the unoptimized grid state to optimal actuator setpoints \cite{khaloieReviewMachineLearning2024b, kotaryEndtoEndConstrainedOptimization2021}, \ie, to approximate the \ac{OPF} with machine learning. By training a neural network, the nonlinear, nonconvex optimization problem is converted into a series of matrix multiplications \cite{wolgastLearningOptimalPower2024}. That is computationally fast, easy to parallelize, and deterministically solvable without convergence issues. The training of neural networks to approximate the \ac{OPF} can be done with all three general machine learning paradigms, \ie, supervised learning \cite{liuNeuralNetworkApproach2024, zhouDeepOPFFTOneDeep2023, panDeepOPFDeepNeural2021, zamzamLearningOptimalSolutions2020a}, unsupervised learning \cite{huangUnsupervisedLearningSolving2024, owerkoUnsupervisedOptimalPower2024, wangFastOptimalPower2023}, and reinforcement learning \cite{wolgastLearningOptimalPower2024, wuConstrainedReinforcementLearning2024, wuChanceConstrainedMDP2024}. Also, combinations into hybrid methods are possible \cite{zhouDeepReinforcementLearning2022, yiRealTimeSequentialSecurityConstrained2024}.
\par 

In this work, we focus on the \ac{RL} approach for learning the \ac{OPF}, which has the advantages of not requiring ground-truth data and strictly separating the general solver algorithm and the domain-specific problem representation \cite{wolgastLearningOptimalPower2024}.
In the \ac{RL} framework, a learning \textit{agent} interacts with its \textit{environment}, which serves as a representation of the problem to solve \cite{suttonReinforcementLearningIntroduction2018}. 
Hence, for the \ac{OPF}, the \ac{RL} environment represents the \ac{OPF} optimization problem with its power grid model, objective function, constraints, and control variables.  
The \ac{RL} agent learns by \textit{observing} the environment (the grid state), performing an \textit{action} (setting control variables), and receiving a \textit{reward} that represents the quality of an action (objective function \& constraints). This procedure gets repeated sequentially until some termination criterion is reached, which marks one \textit{episode}. One \ac{RL} training run consists of thousands or even millions of episodes. 
\par 
Multiple works demonstrated that the design of the environment impacts the \ac{RL} training performance in significant ways \cite{redaLearningLocomoteUnderstanding2020, pengLearningLocomotionSkills2017, ngPolicyInvarianceReward1999, kanervistoActionSpaceShaping2020, zhangStudyOverfittingDeep2018, pardoTimeLimitsReinforcement2018, kimObservationSpaceMatters2021}. In other words, different environment representations of the same problem can improve or reduce the performance of the \ac{RL} algorithm. The \ac{RL}-\ac{OPF} literature consists of various different \ac{OPF} environment design variants, for example, regarding the reward function or the provided observations. However, there is no consensus on which one should be used \cite{wolgastLearningOptimalPower2024}. 
Further, while the literature agrees on the impact of environment design, there exists no general algorithm or methodology to determine the optimal \ac{RL} environment design for a given use case.
\par 
To fill the identified research gaps, we propose a general automated environment design methodology based on multi-objective optimization and the \ac{HPO} framework \cite{bischlHyperparameterOptimizationFoundations2023} and apply it to five different \ac{OPF} problems. Demonstrating the applicability of \ac{HPO} to \ac{RL} environment design is the main contribution of this work. Additionally, we introduce and test environment design options that have not been tested in the \ac{RL}-\ac{OPF} literature so far and derive multiple environment design options that are beneficial for \ac{RL}-\ac{OPF} learning performance, verified by statistical tests.
\par 
Our work is structured as follows: 
After introducing the \ac{OPF} problem in section \ref{sec:background} and discussing the related work regarding RL-OPF and RL environment design in section \ref{sec:related}, we present the \ac{RL} environment design as a multi-objective \ac{HPO} problem in section \ref{sec:methodology}.
In section \ref{sec:opfProblems}, we present the five \ac{OPF} benchmark problems and provide details on our experiments.
In the first results section \ref{sec:performanceResults}, we demonstrate how the environments from the automated design outperform a comparable environment design that was derived manually. 
In section \ref{sec:designResults}, we statistically analyze the optimized environment designs to determine which design decisions were mainly responsible for the improved performance, resulting in general insights on how to design \ac{RL}-\ac{OPF} environments.
In section \ref{sec:verificationResults}, we evaluate the presented methodology and discuss the risk of finetuning the environment to one specific \ac{RL} algorithm. 
We conclude our work with a critical discussion in section \ref{sec:discussion} and a short conclusion in section \ref{sec:conclusion}.

\par 

\section{Background on Optimal Power Flow}\label{sec:background}
The \ac{OPF} refers to a class of optimization problems that incorporate the steady-state power system equations.
In its general form, the \ac{OPF} can be expressed as follows \cite{frankOptimalPowerFlow2012}: 
\begin{equation}
\begin{aligned}
&\mathrm{min} \quad &&J(u, x) \\
&\mathrm{s.t.} \quad &&g(u, x) = 0 \\
& &&h(u, x) \leq 0
\end{aligned}
\end{equation}
where $J$ is the objective function, $g$ represents the equality constraints, $h$ the inequality constraints, $u$ the controllable variables, and $x$ the uncontrollable state variables.
\par 
Typically, the \ac{OPF} constitutes a large-scale, non-linear, and non-convex optimization problem that may include both discrete and continuous control variables. As such, solving it in full generality usually requires a \ac{MINLP} approach.
Due to its broad formulation, the \ac{OPF} framework can be applied to a range of power system optimization tasks, such as economic dispatch, voltage regulation, unit commitment, or topology optimization \cite{frankOptimalPowerFlow2012}. 
\par 
For real-world applications, often more advanced versions of the \ac{OPF} are required. For example, the security-constrained \ac{OPF} incorporates the $N-1$ case by ensuring constraint satisfaction for potential contingencies \cite{capitanescuCriticalReviewRecent2016}. The stochastic \ac{OPF} considers the uncertainty of real-world power systems, for example, by incorporating non-perfect forecasts or the probability of contingencies \cite{faulwasserOptimalPowerFlow2018}. The multi-stage \ac{OPF} optimizes the power flows over multiple sequential time steps, which allows for considering ramp constraints or energy storage systems \cite{faulwasserOptimalPowerFlow2018}. This work mainly focuses on the base case but will consider the advanced \ac{OPF} variants in the discussion if necessary.
\section{Related Work}\label{sec:related}

The following two sections discuss the related work regarding the \ac{RL}-\ac{OPF} and \ac{RL} environment design.

\subsection{Solving the OPF With RL}
We start by discussing the related literature regarding solving the \ac{OPF} with \ac{RL}. Since our work focuses on environment design, which is mainly relevant for \ac{RL}, we will omit the vast literature on supervised and unsupervised learning and refer to the overview by Khaloie \etal \cite{khaloieReviewMachineLearning2024b} instead.
\par 
%
%
%
Liu \etal \cite{liuDeepReinforcementLearning2022} apply the \ac{DDPG} algorithm to a multi-stage \ac{OPF} problem, considering energy storage constraints over multiple time steps. 
Zhen \etal \cite{zhenDesignTestsReinforcementlearningbased2021} combine supervised pre-training with \ac{TD3} and model the \ac{OPF} problem as a 1-step \ac{RL} environment to simplify training.
Nie \etal \cite{nieDeepReinforcementLearning2022} use extensive time-series data to train a \ac{TD3} algorithm to perform voltage control in a microgrid environment.  
%
Zhou \etal \cite{zhouDeepReinforcementLearning2022} use \ac{PPO} with supervised pre-training and convolutional neural networks to solve a stochastic economic dispatch. 
Yizhi Wu \etal \cite{wuChanceConstrainedMDP2024} argue that constraint satisfaction becomes especially challenging in stochastic \ac{OPF} problems and propose chance-constrained \acp{MDP} for stochastic use cases. 
Tong Wu \etal \cite{wuConstrainedReinforcementLearning2024} propose a constrained version of \ac{TD3} and apply it to a stochastic multi-stage economic dispatch. 
Yi \etal \cite{yiRealTimeSequentialSecurityConstrained2024} combine the \ac{SAC} algorithm with various advanced concepts, including pre-training, a linear safety layer, and automatic updates of penalty factors for constraint satisfaction. 
Wolgast and Nieße \cite{wolgastLearningOptimalPower2024} point out the wide variety of different \ac{RL} environment designs in the \ac{RL}-\ac{OPF} literature. They re-implement multiple variants and demonstrate drastic performance differences resulting from the environment design alone.   
\par 
In summary, most of the \ac{RL}-\ac{OPF} works focus on applying increasingly sophisticated \ac{RL} algorithms to more and more complex \ac{OPF} problems. Only \cite{wolgastLearningOptimalPower2024} investigated \ac{OPF} environment design. While showing its impact on performance and deriving first recommendations, they do not provide a general answer on how \ac{RL}-\ac{OPF} environments should be designed. 
Our work aims to fill that gap by proposing an automated environment design methodology and applying it to multiple \ac{OPF} problems. 

\subsection{RL Environment Design}
This section discusses the existing works regarding \ac{RL} environment design.
Peng and van de Panne \cite{pengLearningLocomotionSkills2017} compare different action space representations in robotics tasks. They observe significant performance differences regarding learning speed, final performance, and robustness. 
Kanervisto \etal \cite{kanervistoActionSpaceShaping2020} investigate action space shaping in computer game environments with a focus on discrete actions and removing unnecessary actions.
Kim and Ha \cite{kimObservationSpaceMatters2021} investigate different observation space variants in robotics and also found significant performance differences. 
In an attempt to automate observation space design, they propose a search algorithm to automatically select the best-performing observations. 
Ng \etal \cite{ngPolicyInvarianceReward1999} investigate reward shaping to provide a more useful learning signal. They show multiple examples of how bad reward design can result in faulty policies that exploit the reward function without solving the actual problem. 
Pardo \etal \cite{pardoTimeLimitsReinforcement2018} focus on the episode definition and when a termination signal should be send to the agent. 
Zhang \etal \cite{zhangStudyOverfittingDeep2018} demonstrate how overfitting becomes a problem in \ac{RL} if the learning agent can solve the given problem by memorizing action sequences. 
They recommend strictly separating train and test datasets, as it is common practice in supervised learning but not yet in \ac{RL}.  
Reda \etal \cite{redaLearningLocomoteUnderstanding2020} investigate various environment design decisions in locomotion tasks, including the state distribution, the control frequency, episode termination, action space, etc., and found significant influences on training performance as well. 
\par 
Overall, \ac{RL} environment design is an underexplored topic \cite{redaLearningLocomoteUnderstanding2020, kimObservationSpaceMatters2021, wolgastWhitepaperEnvironmentDesign2024}. While there is consensus that environment design impacts training performance, only Kim and Ha \cite{kimObservationSpaceMatters2021} propose an algorithm for automated observation space optimization.
However, their approach is limited to observations, thus motivating our research on a general and automated environment design methodology.  

\section{Environment Design as Hyperparameter Optimization Problem}\label{sec:methodology}
We can see an \ac{RL} environment as a combination of two parts. First, an immutable underlying
problem definition, which defines the goal, the available observations, and the actions of the \ac{RL} agent. Second, an engineered environment design in the form of the reward function, the observation and action space, the episode definition, and the data distribution, which serve as a representation of the fixed problem to solve. The implementation of these environment design decisions can be chosen freely to maximize performance on the underlying task. \cite{wolgastWhitepaperEnvironmentDesign2024}
\par 
To determine the best-performing environment designs and to enable automation of the
process, we define the \ac{RL}-\ac{OPF} environment design process as an optimization problem.
To represent the two main goals of the \ac{RL}-\ac{OPF} – constraint satisfaction
and optimization performance – we will utilize multi-objective optimization.
The idea is to consider the environment design variables as environment hyperparameters, similar to hyperparameters on the algorithm side, like batch size or learning rate. This allows us to directly apply algorithms and techniques from \acf{HPO}, which is a well-studied field of research \cite{bischlHyperparameterOptimizationFoundations2023, eimerHyperparametersReinforcementLearning2023}, including approaches for multi-objective optimization. 
\par
\ac{HPO} involves finding the optimal hyperparameter setting $\lambda^*$ to maximize model performance, considering a search space $\Lambda$
\begin{equation}
    \lambda^* \in \text{arg min}_{\lambda \in \Lambda} \sim c(\lambda)
\end{equation}
where $c(\lambda)$ is the estimated generalization error of the learner. Usually, this is a black-box optimization problem. Methods range from simple approaches like grid search to sophisticated algorithms like Bayesian optimization \cite{bischlHyperparameterOptimizationFoundations2023}. We transfer this approach to the \ac{RL} environment by defining the environment design parameters as the \ac{HPO} search space $\Lambda$. Consequently, the search space $\Lambda$ contains all potential environment design variables, which can be discrete or continuous, together with their respective ranges, resulting in an $n$-dimensional search space $\Lambda$:
\begin{equation}
    \Lambda = \Lambda_1 \times \Lambda_2 \times ... \times \Lambda_n
\end{equation}
To the best of our knowledge, we are the first to propose using \ac{HPO} for automated environment design. 
\par
\begin{figure}[h]
  \centering
  \includegraphics[width=0.9\linewidth]{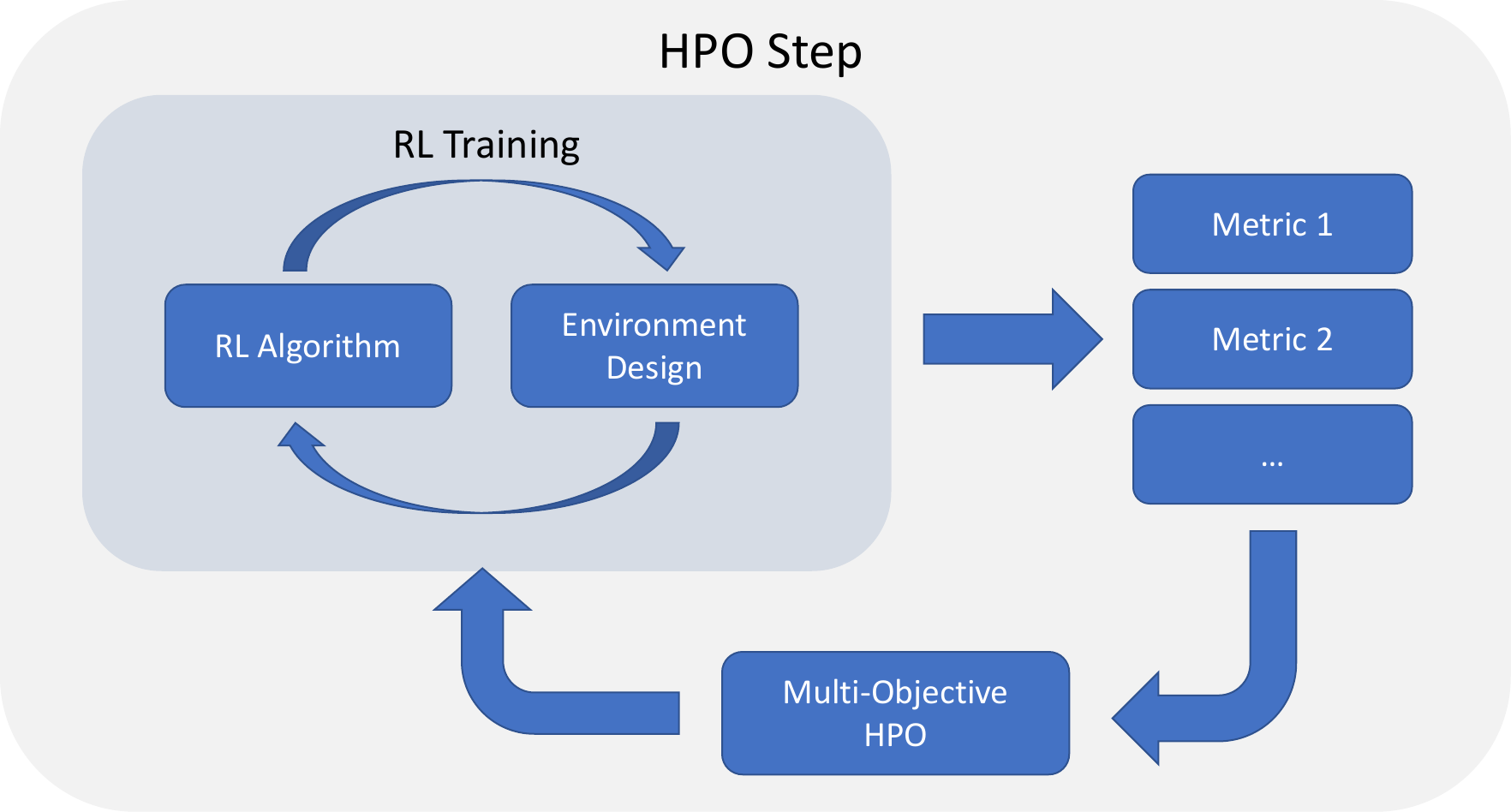}
  \caption{Inner and outer loop of the multi-objective \ac{HPO} for the automated environment design.}
  \Description{TODO}
  \label{fig:process}
\end{figure}

 
Figure \ref{fig:process} visualizes our general approach that consists of an inner and an outer loop. We start with some random initialization of the environment design. In the inner loop, an \ac{RL} algorithm learns on the current environment design. After training, we can evaluate the performance regarding one or multiple metrics, in this case, the performance on the \ac{OPF} task regarding optimization and constraint satisfaction. Based on the performance in the inner loop, some \ac{HPO} algorithm proposes a new environment design in the outer loop. This can be repeated till convergence or until some termination criterion is reached. 
The \ac{RL} algorithm and its hyperparameters stay unchanged during the whole process. 
Note that this general procedure is compatible with almost arbitrary combinations of \ac{RL} algorithm, \ac{RL} problem, and \ac{HPO} optimization algorithm. 

\subsection{Optimization Metrics}
To formulate the automated environment design as an optimization problem, we need to define the metrics to optimize for. For the \ac{OPF}, we need metrics representing two goals: optimization performance and constraint satisfaction. For this, we assume the existence of a baseline conventional \ac{OPF} solver for comparison to evaluate the performance of our approach later on. 
\par 
For the constraint satisfaction performance, we compute the share of invalid solutions when testing on the test dataset. Here, two things have to be considered. First, it should not be evaluated negatively when the agent does not find a valid solution when no valid solution exists. Second, it should be evaluated positively if the agent finds a valid solution where a baseline \ac{OPF} solver fails: 
\begin{equation}
    \Omega = 1 - \frac{N_\text{valid,RL}}{N_\text{valid,base}}
\end{equation}
with the number of valid solutions $N_\text{valid}$. The resulting \invalidShare metric $\Omega$ needs to be minimized and becomes negative if the \ac{RL} agent outperforms the conventional \ac{OPF} solver regarding constraint satisfaction. 
\par 
The optimization performance, or the ability to find the global cost minimum, can be represented by the \meanError $\Delta J$ of the objective values of all valid solutions in comparison with the ground-truth optimal values:
\begin{equation}
    \Delta J_\text{valid} = \frac{1}{N_\text{valid}} \sum_{i=1}^{N_\text{valid}} \left( J_i - J_i^* \right)
\end{equation}
with the objective function $J$.
Only valid solutions are considered here because constraint satisfaction is mandatory, and good optimization performance in invalid states is meaningless. 
Again, the metric needs to be minimized and becomes negative when the \ac{RL} agent outperforms the baseline \ac{OPF} solver. However, while the \invalidShare has an upper bound of one, the \meanError metric has no upper limit. 
With the two metrics \invalidShare and \meanError, the goal is to minimize both metrics, with the point [0, 0] representing equal performance of the \ac{RL} agent and the conventional solver. 

\subsection{Environment Design Space}
We also need to formulate the control variables for optimization and define the search space $\Lambda$ for environment design. In the following, we present \nDecisions \ac{OPF} environment design variables regarding training data, observation space, reward design, episode definition, and action space.    
Most environment design variables are inspired by the \ac{RL}-\ac{OPF} literature and aim to encompass the different variants found in the literature. If applicable, we will explicitly mention the works where a similar variant was used. However, if possible, we avoid discrete variants and aim for continuous design search spaces instead.

\paragraph{Training Data}\label{sec:dataDesign}
In \ac{RL}, the training data distribution is often neglected, which can result in policies that are not transferable to the general case, \ie, overfitting \cite{zhangStudyOverfittingDeep2018}. 
For the \ac{OPF}, to be useful for grid operators or researchers, it is strictly required that the learned policy performs well in real-world scenarios. Therefore, the testing/validation datasets should be as close to reality as possible, which usually means using realistic time-series data of load and generation. Then, the training dataset should be selected such that it results in the best performance on the test dataset \cite{wolgastWhitepaperEnvironmentDesign2024}. In this work, we combine three different training data variants derived from \cite{wolgastLearningOptimalPower2024}:
\par 
The natural choice for the training data is to take the same approach as for the test dataset and sample the environment's state vector $\mathbf{s}$ from some realistic dataset $D$, as done in \cite{nieGeneralRealtimeOPF2019a, nieDeepReinforcementLearning2022, liuDeepReinforcementLearning2022, wolgastApproximatingEnergyMarket2024}:
\begin{equation}
    \mathbf{s} \sim D
\end{equation}
Ideally, that dataset is a subset of the same data as the test data. 
However, the availability of realistic power grid time-series data is limited. Therefore, in an \ac{RL} setting, we can expect repetition of data at some point. Further, time-series datasets will contain a limited number of edge cases, like holidays, extreme weather events, etc. Overall, that might limit the generality of the learned policy \cite{wolgastLearningOptimalPower2024}. 
\par 
To counteract the scarcity of realistic data, we can sample the state vector $\mathbf{s}$ from some random distribution, for example, using a Normal distribution, as done in \cite{yanRealTimeOptimalPower2020},
\begin{equation}
    \mathbf{s} \sim \mathcal{N}(\mathbf{\mu}, \mathbf{\sigma}^2)
\end{equation}
with mean $\mu$ and variance $\sigma^2$, or from a Uniform distribution as done in \cite{zhouDatadrivenMethodFast2020, wooRealTimeOptimalPower2020, zhouDeepReinforcementLearning2022},
\begin{equation}
    \mathbf{s} \sim \mathcal{U}(\mathbf{s}_\mathrm{min}, \mathbf{s}_\mathrm{max})
\end{equation}
with the data range $[\mathbf{s}_\mathrm{min}, \mathbf{s}_\mathrm{max}]$.
Such random data can be created infinitely to create large and diverse datasets. However, most of these randomly sampled grid states will be highly unrealistic \cite{wolgastLearningOptimalPower2024}. Therefore, the \ac{RL} agent might learn a policy for situations that will never happen.
\par 
Using realistic datasets or sampling random data both have their drawbacks and benefits. 
The straightforward approach is to combine both, to sample realistic data for transferability and random data for generality. However, the optimal balancing of randomness and realism in the dataset is not obvious. 
Therefore, we introduce a parameterized sampling method for environment design:
\begin{equation}
    \mathbf{s} \sim 
    \begin{cases} 
        D
        & \text{with probability } x 
        \\
        \mathcal{N}(\mathbf{\mu}, \mathbf{\sigma}^2)
        & \text{with probability } y  
        \\
        \mathcal{U}(\mathbf{s}_\mathrm{min}, \mathbf{s}_\mathrm{max}) 
        & \text{with probability } z \\
        \end{cases} \\
    \quad \text{s.t. } x + y + z = 1.0
\end{equation}
with the respective probabilities $x, y, z$ to sample from each distribution, respectively. These probabilities will be used as parameters for environment design later on. 

\paragraph{Observation Design}\label{sec:obsDesign}
Another important design decision is the choice of the observation space. 
In power grid calculation, each node in the power system has four state properties: active power, reactive power, voltage magnitude, and voltage angle. 
Assuming constant topology, only active power and reactive would be necessary for a power flow calculation, which yields the other variables. The other two can be omitted.
In other words, half of the state properties of electrical power grids are redundant.  
\par 
We can derive two extreme points on how to define the observation space \cite{wolgastLearningOptimalPower2024}. On one hand, we can provide exactly the minimum required observations to the agent as done by \cite{yanRealTimeOptimalPower2020, zhouDatadrivenMethodFast2020, nieDeepReinforcementLearning2022, zhenDesignTestsReinforcementlearningbased2021, liuDeepReinforcementLearning2022, wolgastApproximatingEnergyMarket2024}. These are usually the active and reactive power values of all non-controllable units in the system. On the other hand, we can provide all available system variables as observations \cite{wooRealTimeOptimalPower2020, zhouDeepReinforcementLearning2022, nieGeneralRealtimeOPF2019a}. That might be helpful for constraint satisfaction, considering that exactly these variables are usually constrained in the \ac{OPF}, \eg, the voltage band \cite{wolgastLearningOptimalPower2024}.  
\par 
By comparing the resulting performances of the two extreme points, we neglect the complete space in between. For example, some additional observations may be helpful, while others might even be harmful to performance as shown by Kim and Ha \cite{kimObservationSpaceMatters2021}. 
Due to the large dimensionality of power system states, we will test the performance impact of observations per category. For example, does adding all line loads to the observation space improve performance? This results in five boolean parameters for environment design to define whether to add voltage magnitudes, voltage angles, line loading, transformer loading, and slack bus power values, respectively. This way, we can investigate the space in between the two extreme observation space definitions to determine if adding redundant observations helps to learn the \ac{OPF}.

\paragraph{Reward Design}\label{sec:rewardDesign} 
The reward function $R$ of an \ac{RL}-\ac{OPF} environment must represent the two general goals of minimizing the objective function and satisfying all system constraints, usually by minimizing a penalty function. Again, two general variants can be found in the literature \cite{wolgastLearningOptimalPower2024}: As the first option, we can simply add the penalty function to the objective function as done in \cite{nieGeneralRealtimeOPF2019a, wooRealTimeOptimalPower2020, nieDeepReinforcementLearning2022, liuDeepReinforcementLearning2022, wolgastApproximatingEnergyMarket2024}:
\begin{equation}
    R_\text{sum}(s, a) = -J(s, a) - P(s, a)
\end{equation}
with the objective function $J$, the penalty function $P$, and the action $a$ in state $s$. Alternatively, we can do a case distinction that prioritizes constraint satisfaction over optimization performance as done in \cite{zhouDatadrivenMethodFast2020, zhenDesignTestsReinforcementlearningbased2021, zhouDeepReinforcementLearning2022}:
\begin{equation}
    R_\text{replace}(s, a) =         
    \begin{cases}
        -J(s, a) + r_\mathrm{valid} & \mathrm{if\;valid} \\
        - P(s, a) & \mathrm{else}
    \end{cases}
\end{equation}
with an offset $r_\mathrm{valid}$ to ensure that valid states are always rewarded higher than invalid ones. 
Both reward variants have their pros and cons, with the first prioritizing denser rewards and faster learning and the second one prioritizing constraint satisfaction \cite{wolgastLearningOptimalPower2024}. 
However, these reward functions are the extreme points of a continuous spectrum. To perform automated environment and reward design in this work, we define a parameterizable reward function that encompasses both variants and the full continuous space in between.  
\par 
We start with a weighted sum of two terms that represent the objective function $J$ and the penalty function $P$, respectively.
\begin{equation}
    R_\text{design}(s, a) = (1 - \beta) \hat{J}_\text{norm}(s,a) + \hat{\beta P}_\text{norm}(s,a)
\end{equation}
The \penaltyWeight $\beta$ is the first parameter for environment design. We normalize\footnote{Normalization to mean of zero and a variance of one.} both terms to ensure that potential changes in learning performance can clearly be attributed to the changed characteristics of the function and not its magnitude, which is known to strongly influence learning performance as well \cite{haarnojaSoftActorCriticOffPolicy2018a, hendersonDeepReinforcementLearning2018, vanhasseltLearningValuesMany2016}.
\par 
The adapated objective function $\hat{J}(s,a)$ is defined as follows:
\begin{equation}
    \hat{J}(s,a)_\text{norm} = 
    \begin{cases}
        -J_\mathrm{norm}(s, a) 
        & \mathrm{if\;valid} \\
        -\psi J_\mathrm{norm}(s,a) 
        & \mathrm{else}
    \end{cases}
\end{equation}
The \objectiveShare $0 \leq \psi \leq 1$ is introduced to consider the continuous range between the two extreme points, where $\psi=1$ represents $R_\text{sum}$ and $\psi=0$ represents $R_\text{replace}$. It is another parameter for environment design.
\par 
The adjusted penalty function $\hat{P}(s,a)$ is defined as follows:
\begin{equation}
    \hat{P}(s,a)_\text{norm} = 
    \begin{cases}
        r_\text{valid} 
        & \mathrm{if\;valid} \\
        -P_\text{norm}(s,a) - r_\text{invalid}
        & \mathrm{else}
    \end{cases}   
\end{equation}
The \validReward $r_\text{valid} \geq 0$ represents $R_\text{sum}$ when set to zero and $R_\text{replace}$, if otherwise. The \invalidPenalty $r_\text{invalid} \geq 0$ does the same but punishes invalid states instead of rewarding valid ones \cite{sowerbyDesigningRewardsFast2022}. Both these offset rewards are parameters for environment design.
\par 
Until now, we treated the objective function $J$ for reward calculation as identical to the objective function of the underlying \ac{OPF} problem. 
However, most often, the objective function consists of two parts, where one can be influenced by the control actions while the other cannot. For example, the system losses can be influenced to some extent but cannot be minimized to exactly zero, which results in some fixed offset. As mentioned before, the reward scale influences \ac{RL} training performance significantly. Therefore, an uncontrollable offset in the reward might negatively influence learning performance by superimposing the useful reward signal. 
Hence, we investigate two different objective functions for reward calculation:
\begin{equation}
    J(s,a) = 
    \begin{cases}
        J_\text{OPF}(s,a) - J_\text{init}(s)
        & \text{if \diffObjective} \\
        J_\text{OPF}(s,a)
        & \mathrm{else}   
    \end{cases}
\end{equation}
In the \diffObjective variant, we subtract the estimated uncontrollable part from the objective function. It is estimated by calculating  $J_\text{init}$ of the initial state before acting. The choice between the two options results in another binary parameter for environment design.
\par 
Overall, the reward function for the \ac{OPF} environment is the most complex part regarding environment design with five parameters as degrees of freedom for automated environment design.

\paragraph{Episode Definition}\label{sec:episodeDesign}
The mainly used n-step variant used in \cite{yanRealTimeOptimalPower2020, zhouDatadrivenMethodFast2020, nieDeepReinforcementLearning2022, wooRealTimeOptimalPower2020, zhouDeepReinforcementLearning2022, liuDeepReinforcementLearning2022} formulates the problem as a sequential decision-making problem, \ie, it is formulated as an \ac{MDP} \cite{suttonReinforcementLearningIntroduction2018}. The agent observes the environment, performs an action, receives the reward, observes the resulting state, and so on. The advantage is that the agent can observe unwanted outcomes of its actions and correct them with its following action.
\par
In the 1-step variant \cite{nieGeneralRealtimeOPF2019a, zhenDesignTestsReinforcementlearningbased2021, wolgastApproximatingEnergyMarket2024}, the agent observes the environment's state, performs a single action, receives the reward, and the episode ends. The 1-step variant simplifies the learning because the agent does not need to predict the outcome of its actions over a longer time frame. 
However, the potential drawback is that the agent cannot correct low-performing actions. The 1-step variant equals a contextual bandit, which is a simplified \ac{RL} variant \cite{majzoubiEfficientContextualBandits2020a}.
\par
Regarding the episode definition, we define the \nSteps variable, which is an integer in the range [1, $\infty$], where $1$ represents the 1-step variant and all other options represent the n-step variant.  
\paragraph{Action Space}\label{sec:actionDesign}

One property of power system actuators is that their setpoint range can be limited by dynamic constraints in specific situations. For example, wind turbines and photovoltaic systems have limited feed-in depending on the weather situation, and storage systems have a limited power range when they are full or empty. This becomes a problem when the \ac{RL} agent performs impossible actions like setting photovoltaic feed-in to 100\% at night time. Considering this domain knowledge, the \ac{RL} environment should be designed to prevent such actions because they might sabotage training or can even be exploited by the \ac{RL} agent \cite{wolgastWhitepaperEnvironmentDesign2024}. To deal with such situations, two different action representations are implemented for environment design. We demonstrate this by the example of an active power setpoint $P_\text{set}$:
\begin{equation}
    P_\mathrm{set} = 
    \begin{cases}
    a \cdot (P_\mathrm{max}(s) - P_\mathrm{min}(s)) + P_\mathrm{min}(s) 
    & \text{if \actionAutoscale} \\
    \mathrm{clip}(a \cdot (P_\mathrm{max}^\mathrm{nom} - P_\mathrm{min}^\mathrm{nom}) + P_\mathrm{min}^\mathrm{nom}) 
    & \text{else}
    \end{cases} 
\end{equation}
with the action range $a \in [0, 1]$. In the first case \actionAutoscale, we consider the state-dependent action range [$P_\mathrm{min}(s)$, $P_\mathrm{max}(s)$] for setpoint calculation. The advantage is that the \ac{RL} agent cannot pick an out-of-range action.  
The disadvantage is that the same agent action is interpreted differently depending on the environment state. 
This way, the agent needs to implicitly learn how changes in constraints map to different setpoints.
In contrast, the second variant uses the fixed nominal setpoint range [$P_\mathrm{min}^\mathrm{nom}$, $P_\mathrm{max}^\mathrm{nom}$]. This way, the same action always represents the same setpoint. 
However, this option requires clipping of out-of-range actions. This way, the agent practically utilizes only part of the action space. For example, if we consider a wind turbine that can operate at 50\% power due to low wind, all setpoints $a>=0.5$ will be interpreted the same and yield the same reward. This way, the agent does not receive any feedback on whether, for example, $a=0.6$ or $a=0.8$ is superior, which might result in wasted agent-environment interactions. The \actionAutoscale parameter is another boolean parameter for environment design.

\paragraph{Environment Design Space}\label{sec:designSpace}
Table \ref{tab:envSearchDesignSpace} shows the resulting environment design space with \nDecisions overall design variables.\footnote{We implemented seven more parameters but omitted them for brevity because they yielded no noteworthy results. The respective results can be found in the accompanying repository.} 
If the search space is not naturally bounded, we define the range based on educated guesses and on undocumented pre-studies such that the supposed optimal parameter setting is included in the search space. 

\begin{table}[htp]
    \centering
    \small
    \caption{Environment design search space.}
    \label{tab:envSearchDesignSpace}
    \begin{tabular}{llll}\toprule
        & Design Decision    & Type & Design Space   \\ \midrule
        Reward & \validReward       & float & [0, 2.0] \\
        & \invalidPenalty            & float   & [0, 2.0] \\
        & \objectiveShare    & float & [0.0, 1.0] \\
        & \penaltyWeight              & float & [0.01, 0.99]\\
        & \diffObjective             & boolean & \{True, False\} \\
        \midrule
        Data & \normalShare           & float     & [0\%, 100\%]\textsuperscript{1} \\
        & \uniformShare               & float & [0\%, 100\%]\textsuperscript{1} \\  
        & \simbenchShare             & float  & [0\%, 100\%]\textsuperscript{1} \\
        \midrule 
        Obs & \addVoltageMag     & boolean            & \{True, False\}\\
        & \addVoltageAngle             & boolean    & \{True, False\}  \\
        & \addLineLoad              & boolean   & \{True, False\} \\
        & \addTrafoLoad             & boolean    & \{True, False\} \\
        & \addSlackPower          & boolean       & \{True, False\}\\
        \midrule
        Episode & \nSteps                   & integer & \{1, 3, 5\} \textsuperscript{2} \\ 
        \midrule
        Action & \actionAutoscale & boolean  & \{True, False\} \\ 
        \bottomrule
        \addlinespace
        \multicolumn{3}{l}{\textsuperscript{1} Constrained to a total of 100\%.} \\
        \multicolumn{3}{l}{\textsuperscript{2} Restricted to \{1\} after initial experiments.} \\
    \end{tabular}
\end{table}

The first test experiments already clearly demonstrated that \nSteps $>1$ results in drastically worse performance for all environments, which is why the search space was restricted to $\{1\}$ only. That was necessary to ensure that this single parameter does not dominate the whole evaluation and all other design variables.


\section{Integrating the OPF problems}\label{sec:opfProblems}
After defining the environment design problem as multi-objective HPO problem, we will now define the relevant OPF problems, and show the experimental setup used for the evaluation.
\subsection{Optimal Power Flow Use Cases}\label{sec:benchmarkProblems}
For the evaluation of our methodology, we utilize five benchmark environments from the \opfgym\footnote{\url{https://github.com/Digitalized-Energy-Systems/opfgym}} framework \cite{Wolgast_OPF-Gym}, first introduced in \cite{wolgastLearningOptimalPower2024}. 
By using an open-source benchmark framework, we ensure reproducibility of our experiments and comparability with other research. As discussed before, we will consider the \ac{OPF} problem to be fixed while seeing the environment design as open for optimization. Hence, we will deviate from the default environment design in \opfgym.
\par 
The following sections concisely describe the utilized benchmark environments. They represent different variants of the \ac{OPF} problem. For example, active and reactive power setpoints are considered as control variables, market- and non-market-based \ac{OPF}, different actuators like generators, loads, and storage systems, and so on. 
For a comprehensive description of the environments and their underlying \ac{OPF} problems, refer to the \opfgym documentation\footnote{\url{https://opf-gym.readthedocs.io/en/latest/benchmarks.html}}.
\par 
All five environments share the same general constraints: They are subject to
voltage band constraints with the voltage $V$ of all buses $B$, 
\begin{equation}
   V_b^\text{min} \leq V_b \leq V_b^\text{max} \; \forall \; b \; \in \; B,
\end{equation}
overload constraints of all lines $L$ and transformers $T$ with the apparent power flow $S$, 
\begin{equation}
    S_l \leq S_l^\text{max} \; \forall \; l \; \in \; L, 
\end{equation}
\begin{equation}
    S_t \leq S_t^\text{max} \; \forall \; t \; \in \; T,
\end{equation}
and limited active and reactive power ranges $[P_s^\text{min}, P_s^\text{max}]$ and $[Q_s^\text{min}, Q_s^\text{max}]$ of the slack bus $s$. 
\begin{equation}
   P_s^\text{min} \leq P_s \leq P_s^\text{max}
\end{equation}
\begin{equation}
   Q_s^\text{min} \leq Q_s \leq Q_s^\text{max}
\end{equation}
Additionally, the power balance equations and the active/reactive power setpoint ranges of the actuators are inherently considered by \opfgym, which is why we do not discuss them here explicitly. For each environment, some constraints are more relevant than others, which will be mentioned in the respective cases.

\subsubsection{Voltage Control}
The \voltageControl environment is an optimal reactive power flow problem, where generators and storage systems are controlled to minimize system-wide active power losses $P_\text{loss}$ subject to all constraints with a focus on the voltage band constraints and the slack reactive power constraint. It has a 14-dimensional continuous action space, which results from ten controllable generators and four controllable storage systems.
\begin{equation}
    \text{min} \; J = P_\text{loss}
\end{equation}

\subsubsection{\loadShedding}
The \loadShedding environment aims for constraint satisfaction with cost-minimal load shedding in a commercial area. The load shedding prices $p_a$ of the actuators $a$ are randomly sampled for different states $s$ to consider different state-dependent preferences of the load owners in an ancillary service market. Additionally, multiple energy storage systems can be controlled. Here, especially the line load constraint and the slack power flow constraint are relevant. The environment has a 16-dimensional continuous action space from 15 controllable generators and one storage system.   
\begin{equation}
    \text{min} \; J = \sum_{a \in A} P_a \cdot p_a(s)
\end{equation}

\subsubsection{Economic Dispatch}
The \ecoDispatch is the most relevant \ac{OPF} variant. The objective is to satisfy load demand with cost-minimal active power generation subject to multiple constraints, from which the line load constraint is most prevalent. 
The environment has 42 generators, resulting in a 42-dimensional continuous action space.
\begin{equation}
    \text{min} \; J = \sum_{a \in A} P_a \cdot p_a^P(s)
\end{equation}

\subsubsection{Reactive Power Market}
The \qmarket environment is a market-based variant of the \voltageControl environment. It is an optimal reactive power flow with priced reactive power setpoints of generators. The objective is to minimize the sum of loss costs and reactive power market costs for the grid operator. The most relevant constraints are the voltage band and the restricted slack reactive power flow, which enforces local reactive power procurement. It has 10 controllable generators and therefore a 10-dimensional continuous action space.
\begin{equation}
    \text{min} \; J = P_\text{loss} \cdot p_\text{loss}^P + \sum_{a \in A} Q_a \cdot p_a^Q(s)
\end{equation}

\subsubsection{Maximize Renewables}
In the \maxRenewable environment, the objective is to maximize overall active power feed-in of all controllable renewable generators $G$. The available actuators are the active power setpoints of the renewable generators and some storage systems. The storage systems are not part of the objective function but can be used for constraint satisfaction, where the voltage band, the line load, and the trafo load are relevant. The environment's continuous action space is 18-dimensional with 15 generators and three storages.
\begin{equation}
    \text{min} \; J = -\sum_{g \in G} P_g
\end{equation}

\subsection{Experiments}

The experiments for this work are performed as follows. For multi-objective \ac{HPO} in the outer loop, we utilize the open-source framework \optuna \cite{akibaOptunaNextgenerationHyperparameter2019}. As the optimization algorithm, we choose the \textit{NSGAIIISampler} \cite{debEvolutionaryManyObjectiveOptimization2014} with its default parameters. It was chosen because it outperformed other optimizers in undocumented pre-studies and is capable of multi-objective optimization. Overall, 100 outer loop \ac{HPO} steps are performed, where each optimization step represents one environment design setting.
\par 
Regarding the inner \ac{RL} loop, one important aspect to consider is the stochasticity of \ac{RL} experiments, which might distort evaluation with positive or negative outliers \cite{eimerHyperparametersReinforcementLearning2023}. To counteract stochasticity, each sample consists of three training runs with different seeds to achieve a robust performance estimation. The metrics are averaged over the three runs, respectively. Each single \ac{RL} training run is performed with the basic \ac{RL} algorithm \ac{DDPG} \cite{lillicrapContinuousControlDeep2016} for 40k training steps. 
The short training time of 40k steps\footnote{For comparison, in \cite{wolgastLearningOptimalPower2024}, the training times were 1-2 million steps for very similar environments.} was chosen to favor fast-converging environments and to make the problem computationally tractable. We will later test the validity of that decision.
The off-policy \ac{DDPG} algorithm was chosen over more state-of-the-art \ac{PPO} \cite{schulmanProximalPolicyOptimization2017} or \ac{SAC} \cite{haarnojaSoftActorCriticOffPolicy2018a} because we found it to be converging faster than both algorithms in an undocumented pre-study. That aspect is especially important here, considering the short training times discussed before. 
The \ac{DDPG} hyperparameters can be found in Table \ref{tab:ddpgHPs} in the Appendix. We will test later if the results are transferable to other \ac{RL} algorithms.
\par 
The accompanying environment's time-series datasets with 35k data points are split into training, validation, and testing datasets using randomized nested resampling as described by Bischl \etal \cite{bischlHyperparameterOptimizationFoundations2023}. First, the data is split deterministically into 80\% training data and 20\% testing data. The testing data is neither used for training nor for environment design evaluation during optimization. That is to prevent a positive bias by picking environment designs that perform best on the test dataset by chance \cite{bischlHyperparameterOptimizationFoundations2023}. Instead, they will be used for verification of results later on. After the test split, the remaining 80\% are randomly split into training and validation datasets. For this work, we use 7k samples for training. This small amount is explicitly chosen to create an artificial shortage of \simbenchShare to determine if randomly sampled data can compensate for that lack of data, as discussed in section \ref{sec:dataDesign}. This allows us to evaluate if random data sampling method can replace realistic data without a performance drop.
\par 
The performance evaluation during environment design happens on the validation data. Again, we have to consider stochasticity. To achieve a robust performance estimation, we perform four overall evaluations on the validation data after  25k, 30k, 35k, and 40k training steps, respectively. This procedure dampens outliers and prefers quick learning over slow learning.

\subsection{Baseline Environment Design}
The previously described approach of multi-objective optimization yields a Pareto-front of non-dominated environment designs. To determine if the optimized environment designs result in higher performance than a manual design, we consider a manually derived environment design as a baseline for comparison. 
For that, we use the environment design derived by Wolgast and Nieße \cite{wolgastLearningOptimalPower2024} because it is the most comprehensive empirical analysis of \ac{RL}-\ac{OPF} environment designs, as discussed in section \ref{sec:related}. They re-implemented the existing options from various \ac{RL}-\ac{OPF} publications and compared them regarding performance. Hence, we assume that their resulting design is the best-performing \ac{RL}-\ac{OPF} environment design so far. Further, the analysis of \cite{wolgastLearningOptimalPower2024} was done on two of the \opfgym environments, which improves comparability to our work, considering that we also use the \opfgym benchmark. 
\par 
The exact baseline environment design can be found in Table \ref{tab:baseDesign} in the Appendix. It uses no offset rewards, no random training data, no redundant observations, 1-step episodes, and action autoscaling.
However, we deviate from that environment design in one aspect by using a normalized reward instead of an unscaled reward to ensure comparability (compare section \ref{sec:rewardDesign}). Further, they do not provide any recommendation regarding the weighting of the penalty vs. the objective function, which is why we consider multiple values for the \penaltyWeight. We use the weights $\{0.1, 0.3, 0.5, 0.7, 0.9\}$. The training runs with the manual design are performed the same as described in the previous section, except using ten different seeds to prevent outliers.

\section{Performance Evaluation}\label{sec:performanceResults}

The following sections compare the resulting performance of the proposed \ac{HPO}-based automated design with the manually derived baseline design. 
\par 
\par 
\subsection{Economic Dispatch}
\begin{figure}[ht]
  \centering
  \includegraphics[width=0.99\linewidth]{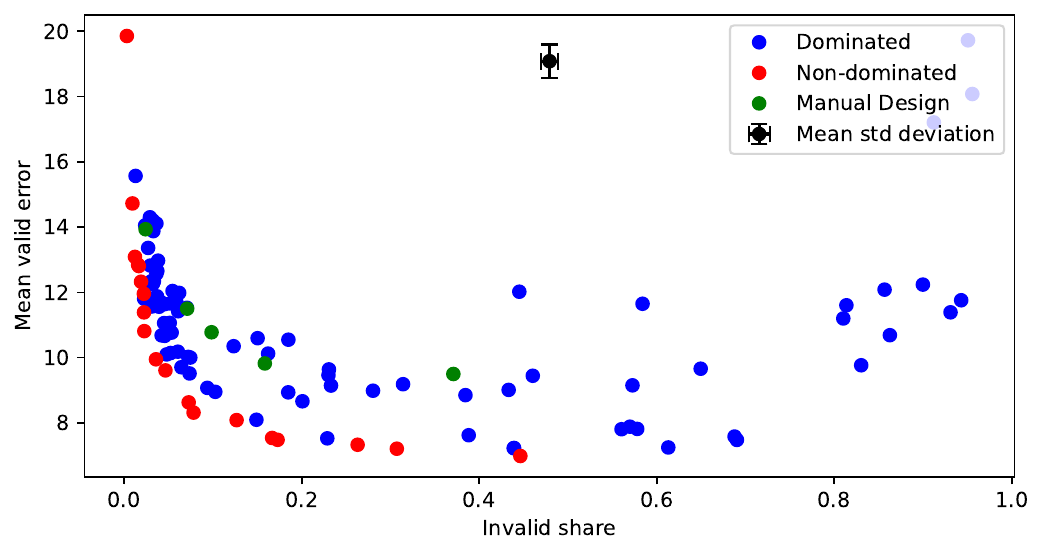}
  \caption{Pareto-front and distribution of samples for the \ecoDispatch environment, including the baseline results.}
  \Description{TODO}
  \label{fig:eco}
\end{figure}

Figure \ref{fig:eco} shows the results for the \ecoDispatch environment.
The figure contains all non-dominated and dominated solutions from the multi-objective optimization in red and blue, respectively. Additionally, we added the baseline runs with the manual design in green. The cross in the upper Figure shows the mean standard deviation calculated over all 100 samples of the \ac{HPO}. 
We can observe a strong tradeoff between the two metrics, which results in the typical curved Pareto-front. All points from the manual design get outperformed on the right above the Pareto-front, which shows that the manual designs get strictly dominated by the solutions from automated design. That is especially true for the samples with worse constraint satisfaction (lower \penaltyWeight). 
Altogether, the solutions from the automated design significantly outperform the manual design. 
\subsection{Load Shedding Environment}
\begin{figure}[ht]
  \centering
  \includegraphics[width=0.97\linewidth]{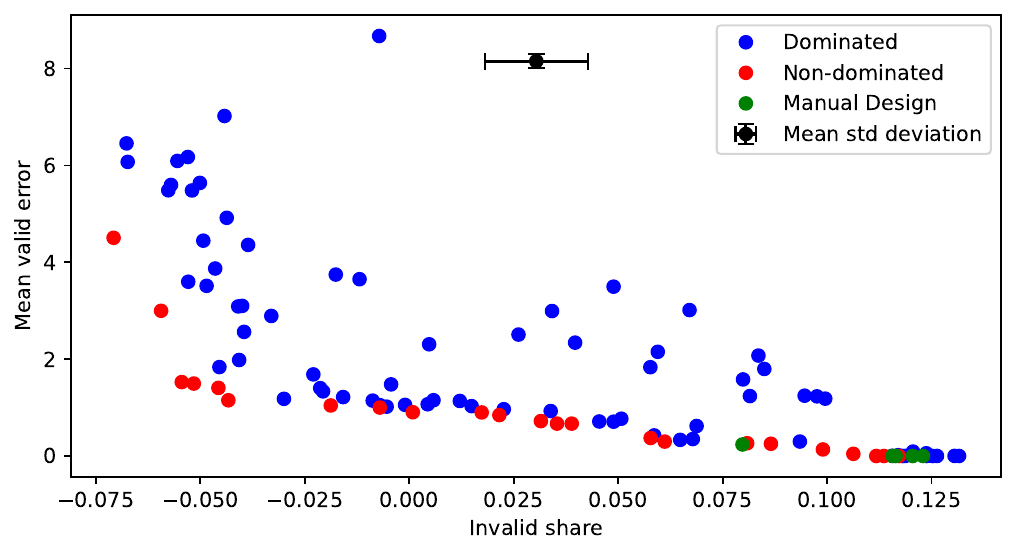}
  \caption{Pareto-front and distribution of samples for the \loadShedding environment, including the baseline results.}
  \Description{TODO}
  \label{fig:load}
\end{figure}

Figure \ref{fig:load} shows the resulting distribution of training performances for the \loadShedding environment. 
Again, the non-dominated solutions of the automated design result in a curved Pareto-front that visualizes the trade-off between constraint satisfaction and optimization performance.
However, in this case, the automated designs do not strictly outperform the manual design. Instead, all manual designs are located on the far right end of the Pareto-front, almost independent of the chosen penalty weight. Therefore, the environment design from \cite{wolgastLearningOptimalPower2024} is generally competitive. However, it also shows that even with varying penalty weights, only a very small part of the search space was considered. While the manual design consistently failed to achieve constraint satisfaction, the left-most solutions from the automated design demonstrate that the \ac{RL} agent can even outperform the conventional solver by 7\% regarding constraint satisfaction. Without the explorative character of the multi-objective environment design, that would not have been possible. 

\subsection{Remaining Environments}
For conciseness and to prevent repetitions, we will summarize the results for the remaining three environments, \voltageControl, \maxRenewable, and \qmarket. Their respective Pareto-fronts can be found in Appendix \ref{app:results}. 
For all three, the non-dominated solutions from the automated design either outperform the solutions from the manual design or achieve the same performance. In the \voltageControl environment, the automated design even outperforms the manual one by multiple standard deviations.

\section{Environment Design Evaluation}\label{sec:designResults}

One characteristic of multi-objective optimization is that we cannot extract a single best environment design from the previous results Instead, we receive a set of non-dominated solutions as shown before. However, we can draw general conclusions about which environment design decisions are relevant for which metric and potentially also which environment design decisions are generally better than others.
\par 
To do this, we split the generated solutions into two groups – dominated and non-dominated solutions – and then test if there are statistically significant differences regarding the environment design. For example, if all non-dominated solutions contain the voltage magnitude in the observation space, while the distribution in the dominated set is 50/50, there is a high probability that voltage magnitude observations are required for generating non-dominated environment designs.
\par 
While splitting regarding dominated/non-dominated is natural for multi-objective optimization, other criteria are possible as well. In the following, we will split the generated solutions regarding four criteria: 
\begin{enumerate}
    \item \pareto: Non-dominated vs. dominated
    \item \constraints: Good \invalidShare vs. bad \invalidShare
    \item \objective: Good \meanError vs. bad \meanError
    \item \utopia: Sum of both normalized metrics.  
\end{enumerate}
Regarding the latter three, we will perform the split by putting the top 20\% solutions into one group and the bottom 80\% into the other group.\footnote{We chose the 20/80-split based on an undocumented sensitivity analysis. That analysis showed that the choice of the cut-off point influences the general results only marginally.} Then, we determine the p-value by using Welch's t-test \cite{welchGeneralizationSTUDENTSproblemWhen1947} for the continuous parameters and the chi-squared test \cite{pearsonCriterionThatGiven1900} for the discrete ones. We reject the null hypothesis that a design decision
does not impact performance if $p < 0.05$.
This way, we can test all environment design decisions for statistical significance regarding the four evaluation criteria.
\par 
This general procedure can be performed for individual environments, but also for all environments together.
Since the focus of our work is on the general methodology, we will first discuss the overall results, followed by one specific design by the example of the \loadShedding environment.\footnote{The statistically significant environment design decisions for the other individual environments can be found in Appendix \ref{app:results}.}  
\par 
\subsection{General OPF Environment Design}
This section discusses the design decisions that have a statistically significant influence over all 500 samples. For that, we extract the top 20\% designs from all five environments, respectively, as described before. 
Additionally, for the \pareto criterion, we also have to consider that the environments have different numbers of non-dominated solutions. We use Fisher's method \cite{fisherStatisticalMethodsResearch1992} to combine the respective p-values to not favor the environments with more non-dominated solutions. 
\par
\begin{figure}[htb]
  \centering
  \includegraphics[width=0.99\linewidth]{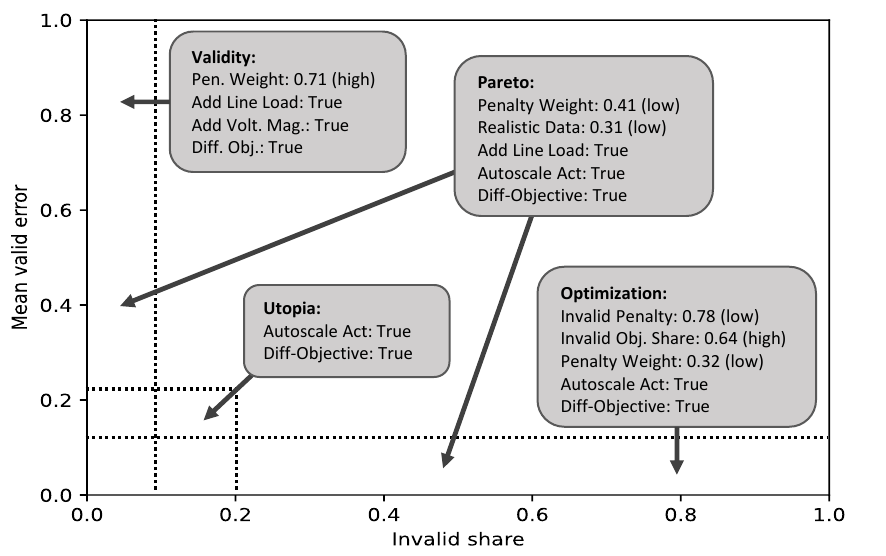}
  \caption{Statistically significant environment design decisions over all five environments for all four evaluation criteria, respectively.}
  \Description{TODO}
  \label{fig:designOverallAnnotated}
\end{figure}

 
Figure \ref{fig:designOverallAnnotated} shows the environment design decisions with statistically significant influence on the performance regarding the four evaluation criteria over all five environments. For the discrete design decisions, we show which exact variant is especially prevalent in the high-performing group. For the continuous variables, we show the mean of the top group with a comment if it is high or low relative to the search space.
Note that we removed the actual data points for a better overview. The dotted lines hint at the respective areas of the evaluation criteria in a stylized manner.
\par 
The first noteworthy result is that while we investigated \nDecisions different design decisions, only 2-5 of them resulted in statistically significant performance impacts over all five environments. The same is true for each singular environment (see Appendix \ref{app:results}). In conclusion, while environment design impacts performance on a general level, the same is not true for each individual design decision. Some are far more important than others. 
\par 
Over all five environments and all evaluation criteria, only some design decisions are noteworthy to discuss. 
A high \penaltyWeight benefits constraint satisfaction but harms optimization performance. However, we cannot derive novel insights from that observation because the general trade-off is already known \cite{wolgastLearningOptimalPower2024, zhangEvaluatingModelFreeReinforcement2023, khaloieReviewMachineLearning2024b}. Instead, this observation can be seen as a confirmation that the results align with the existing knowledge.
\par 
Using the \diffObjective and \actionAutoscale options seems to consistently benefit performance, almost independent of the evaluation criterion. Considering that these results were generated over five different environments and 1500 training runs, we can conclude that both should be considered for designing \ac{OPF} environments.
\par 
Also noteworthy is that a relatively low \simbenchShare share resulted in significantly more non-dominated Pareto-front solutions, while the \normalShare and \uniformShare shares have not resulted in any statistically significant results. This confirms that time-series data can indeed be supplemented by randomly sampled data to some extent. That is in contrast to the results from Wolgast and Nieße \cite{wolgastLearningOptimalPower2024}, which suggested that randomly sampled data cannot result in competitive performance. Instead, our results show that, assuming a limited existing dataset, adding randomly generated data may improve performance. However, our results do not provide information if a uniform or a normal distribution is better for random sampling. It is probably a use-case-specific choice. For example, Figure \ref{fig:eco} in the Appendix indicates that for the \ecoDispatch environment, a high \uniformShare share is beneficial, while Figure \ref{fig:qmarketAnnotated} suggests the opposite for the \qmarket environment.
\par 
Again in contrast to \cite{wolgastLearningOptimalPower2024}, the results indicate that adding redundant observations can improve performance, especially regarding constraint satisfaction. That is the case for the line loading observations and the voltage magnitudes. However, considering that in different \ac{OPF} variants, different constraints are more important than others, we can assume again that it is use-case-specific.

\subsection{Specific Environment Design}

\begin{figure}[htb]
  \centering
  \includegraphics[width=1.0\linewidth]{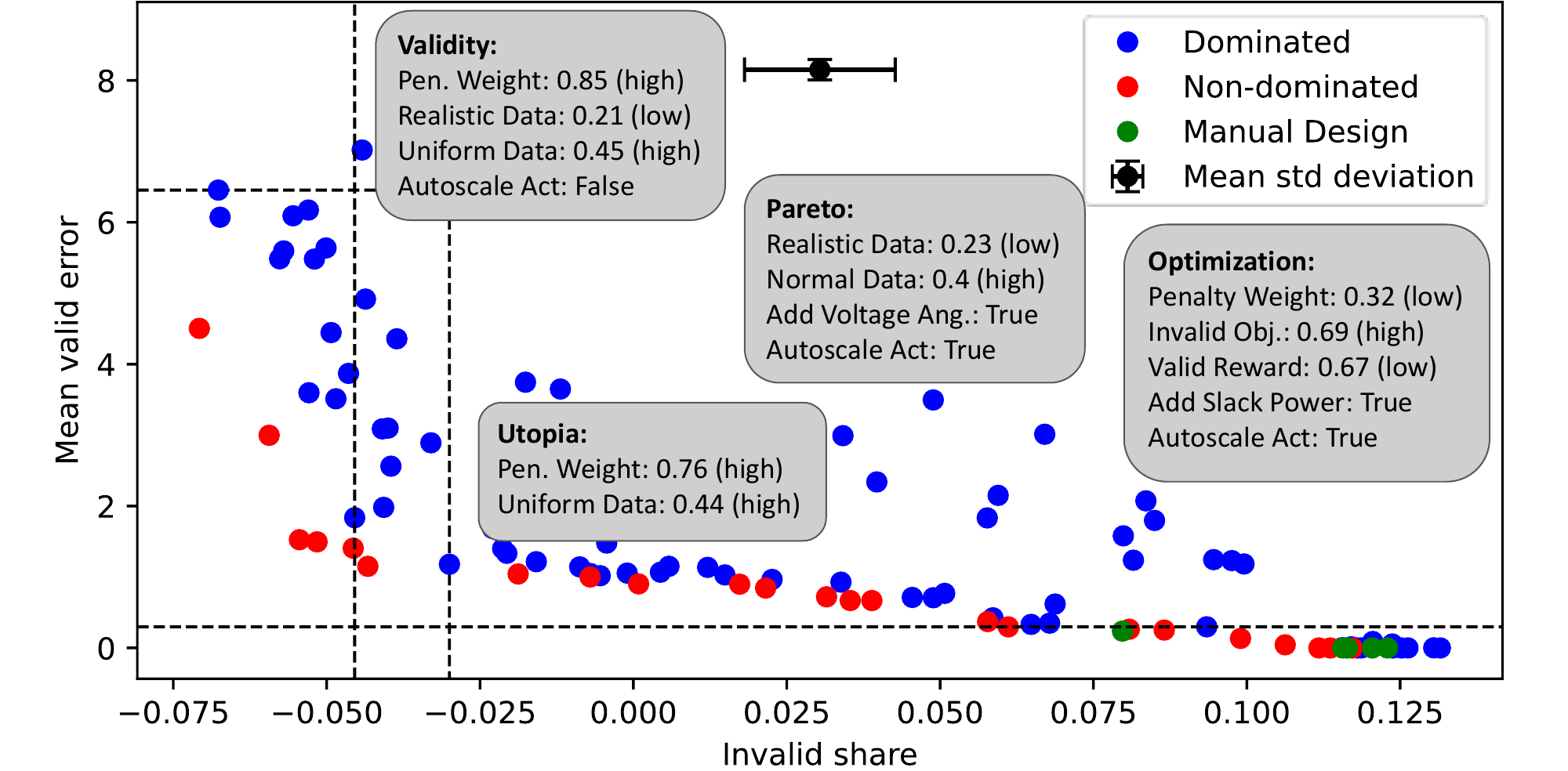}
  \caption{\loadShedding Pareto-front and distribution of results, including all statistically significant environment design decisions.}
  \Description{TODO}
  \label{fig:designLoadAnnotated}
\end{figure}

This section performs the same analysis for the \loadShedding environment. Figure \ref{fig:load} demonstrated that the baseline failed to achieve competitive constraint satisfaction performance. Figure \ref{fig:designLoadAnnotated} extends the illustration with the statistically significant design decisions for each evaluation criterion again. 
\par 
Some of the results are expected again. For example, a low \validReward is good for optimization performance, while a high \penaltyWeight is required for strong \constraints performance. However, the baseline design failed with constraint satisfaction, although it used a high \penaltyWeight, too. Hence, we can conclude that the training data distribution is the determining factor for the superior \constraints performance of the automated design in this environment. Figure \ref{fig:loadAnnotated} shows that a low \simbenchShare share and high \normalShare or \uniformShare share are beneficial. This strongly suggests that a higher share of randomly sampled states enables the \ac{RL} agent to learn better constraint satisfaction in this environment.
\par 
Besides the emphasis on randomly sampled states, the \loadShedding environment deviates from the general results in Figure \ref{fig:designOverallAnnotated} in other aspects as well. For example, the  \diffObjective reward or the \actionAutoscale do not appear to have the same clear positive effects on performance. 

Altogether, we demonstrated how statistical tests can be used to determine design decisions that impact performance in statistically significant ways.
However, note that while we could not show statistical significance for some other design decisions, that does not mean that they do not influence performance at all. It is also possible that more samples would have been required to show the statistical significance of their influence.

\section{Verification of Optimized Designs}\label{sec:verificationResults}
In the previous experiments, we made multiple decisions that might have resulted in a positive bias.
To make the automated design computationally tractable, we chose short training times of 40k steps, which assumes that the resulting environment design also works well for longer training times. We chose \ac{DDPG} for environment design because it performed well in trial runs, but it remains unclear if the environments that were optimized for \ac{DDPG} also perform well with other \ac{RL} algorithms. Further, we have not yet extracted specific environment designs from the samples. Can we extract environment designs from the results that reproducibly dominate the baseline?
\par
\begin{figure*}[t]
    \centering

    \begin{subfigure}[t]{0.45\textwidth}
        \centering
        \includegraphics[width=0.93\textwidth]{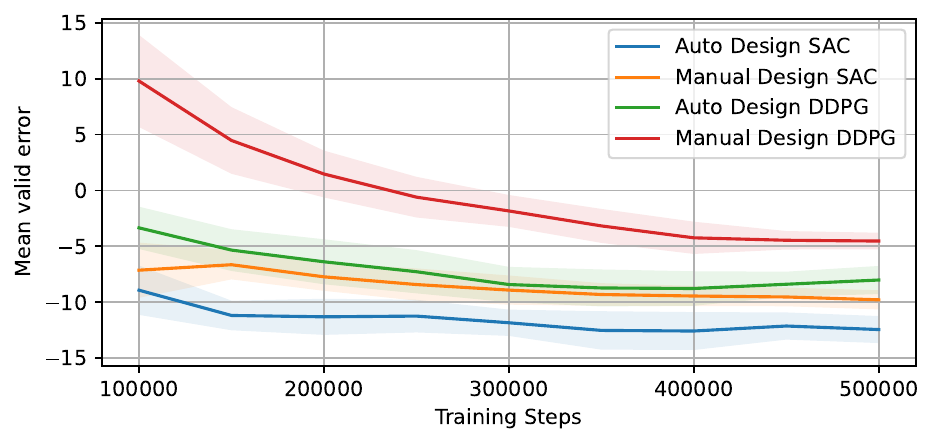}
        \caption{\voltageControl with the \meanError metric.}
        \label{fig:subfig1}
    \end{subfigure}
    \hfill
    \begin{subfigure}[t]{0.45\textwidth}
        \centering
        \includegraphics[width=0.93\textwidth]{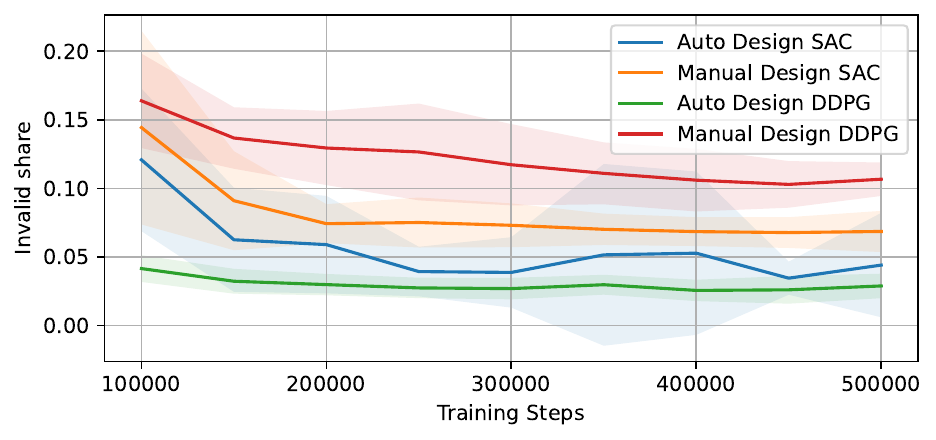}
        \caption{\voltageControl with the \invalidShare metric.}
        \label{fig:subfig2}
    \end{subfigure}

    \vspace{0.3cm}
    \begin{subfigure}[t]{0.45\textwidth}
        \centering
        \includegraphics[width=0.93\textwidth]{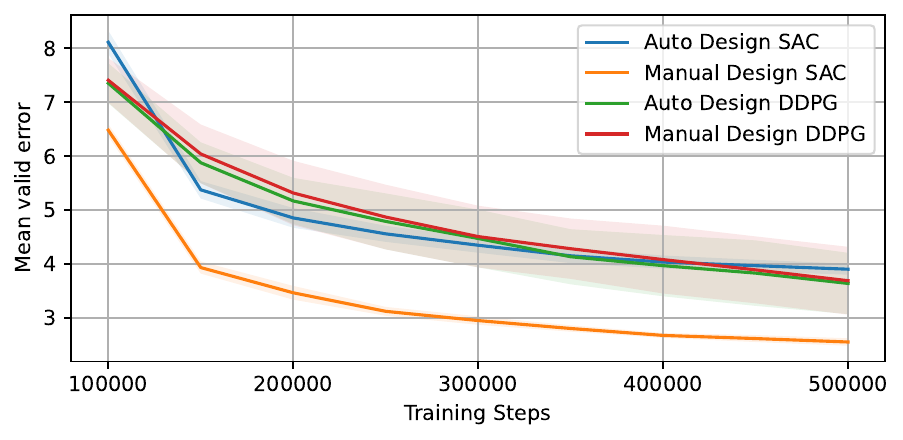}
        \caption{\ecoDispatch with the \meanError metric.}
        \label{fig:subfig3}
    \end{subfigure}
    \hfill
    \begin{subfigure}[t]{0.45\textwidth}
        \centering
        \includegraphics[width=0.93\textwidth]{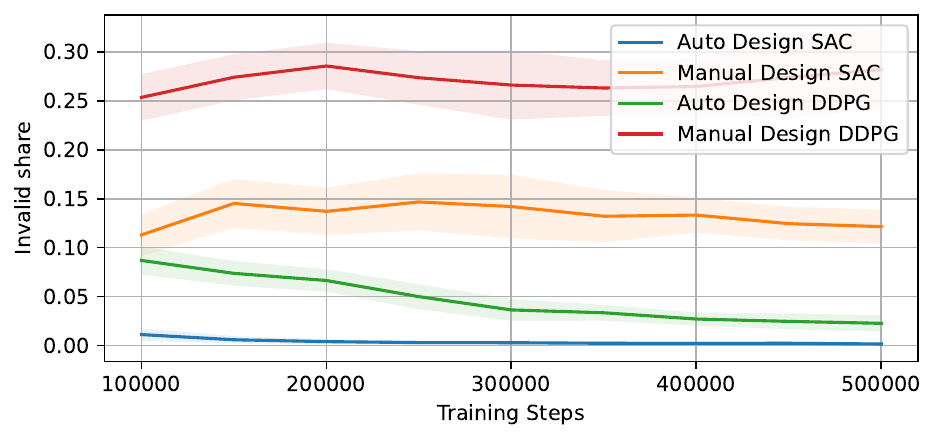}
        \caption{\ecoDispatch with the \invalidShare metric.}
        \label{fig:subfig4}
    \end{subfigure}
    
    \caption{Learning curves for both verification environments, \ecoDispatch and \voltageControl, for both considered performance metrics and for both the \ac{DDPG} and the \ac{SAC} algorithm. The colored areas mark one standard deviation. All plots are created with a rolling average over two steps.}
    \label{fig:verification}
\end{figure*}

To verify the performance gains from the automated environment design, we will perform verification training runs.
For that, we will focus on the two environments where the automated design clearly outperformed the baseline design regarding both metrics: \ecoDispatch and \voltageControl. The hypothesis is that we can extract environment designs from the previous optimization that will result in reproducibly superior performance compared to the baseline design.
\begin{enumerate}
    \item We increase the training time from 40k steps to 500k. Further, we now use the full available training set instead of the reduced dataset used before.
    \item We perform the experiments for \ac{DDPG}, which was used for the optimization, and \ac{SAC} \cite{haarnojaSoftActorCriticOffPolicy2018a}, which is considered to be the state-of-the-art off-policy \ac{RL} algorithm.
    \item For both environments, we pick a combination of the best five samples regarding the \utopia criterion to prevent outliers. For the continuous variables, we use the mean; for the discrete ones, we take the most used setting. We do that for all design decisions, regardless of statistical significance. The resulting exact environment designs can be found in Table \ref{tab:baseDesign} in the Appendix. 
\end{enumerate}
Further, we now compute the performance metrics on the test datasets, which were not used for automated design before, as it is good practice in \ac{HPO} \cite{bischlHyperparameterOptimizationFoundations2023}. 
For the baseline design, we picked the penalty weight that resulted in the best \utopia performance in the previous training runs. That is a penalty weight of $0.5$ for \ecoDispatch and $0.1$ for \voltageControl.
All metrics are calculated as the mean of 10 different random seeds to consider the stochasticity of results \cite{eimerHyperparametersReinforcementLearning2023, hendersonDeepReinforcementLearning2018}.
The resulting training curves are shown in Figure \ref{fig:verification}. The 2x2 subfigures show the performance regarding the two metrics \meanError and \invalidShare for the two environments, respectively.
\par 
Regarding the \meanError metric in the \voltageControl environment, the automated design outperforms the manual design for both \ac{RL} algorithms.
Regarding the \invalidShare metric, again, the automated design generally outperforms the baseline for both algorithms. However, the combination of automated design plus \ac{SAC} resulted in multiple outliers, which can be deduced from the high standard deviation.
\par 
For the \ecoDispatch environment, again, the automated design outperforms the manual design, however, with one exception. 
While the automated design plus \ac{SAC} resulted in almost perfect constraint satisfaction, the optimization performance was the worst out of the four combinations.
Hence, this is the only case where the automated design does not dominate the manual design regarding both metrics. However, it also does not get dominated by the baseline.
The switch from \ac{DDPG} to \ac{SAC} resulted in a stronger focus on constraint satisfaction for the \ecoDispatch environment. The exact reason for this behavior is out-of-scope for this work.
Therefore, for this combination, we cannot draw a general conclusion that one environment design is superior to the other.
\par 
Altogether, we can conclude that the optimized environment designs from the multi-objective optimization reproducibly outperform and dominate the baseline design, if we use the same \ac{RL} algorithm. However, our results also suggest that the optimized environment design is not fully transferable from one \ac{RL} algorithm to the other, which indicates some kind of overadjustment of the environment to the \ac{RL} algorithm, similar to overfitting \cite{zhangStudyOverfittingDeep2018}. 

\section{Discussion}\label{sec:discussion}
The results of section \ref{sec:performanceResults} and \ref{sec:verificationResults} demonstrate that our methodology for automated environment design reproducibly results in environments that outperform manually designed environments. In none of the five use cases, the manual design achieved performances below the Pareto-front of the automated design. Besides the performance gains, multi-objective optimization also allows us to extract environment designs from arbitrary points of the Pareto-front. For our verification, we picked an environment design that balances both metrics, but it would also have been possible to focus completely on constraint satisfaction or optimization performance, depending on the specific use case. 
\par 
Section \ref{sec:designResults} demonstrated how statistical tests can be used to determine the most crucial environment design decisions. The results over five \ac{OPF} environments demonstrate how this methodology can create novel knowledge for the general case. Our findings on the usefulness of the \diffObjective, the \actionAutoscale, and the possibility of mixing randomly generated data with time-series data are especially noteworthy since they have not been discussed in the \ac{RL}-\ac{OPF} literature yet. These results will provide a valuable basis for future \ac{RL}-\ac{OPF} research.
\par 
However, we should be careful not to draw too general conclusions. While we showed statistical significance for five different \acp{OPF}, there is a wide range of \ac{OPF} use cases that we did not explore, like the multi-stage or the stochastic \ac{OPF}. For example, the 1-step environment design outperformed the n-step design. However, that result is most probably not transferable to the multi-stage \ac{OPF}, which strictly requires sequential decision-making. 
\par 
While we showed the general usefulness of our methodology for automated environment design, there are also some drawbacks to our approach.
Our verification results in section \ref{sec:verificationResults} suggest that an environment design that was performed for one \ac{RL} algorithm is not necessarily transferable to other algorithms. One important advantage of the \ac{RL} framework is its modularity that comes from the agent/environment-split. If the performance drops by switching to another \ac{RL} algorithm, that advantage no longer exists. Making our automated environment design robust to subsequent algorithm changes, is an important step for future work.
\par

\par 
While reusing the \ac{HPO} framework and algorithms for environment design results in reusability of existing algorithms and methods, it also inherits the drawbacks of \ac{HPO}. The most relevant one is the required computational effort \cite{bischlHyperparameterOptimizationFoundations2023}. Overall, 300 \ac{RL} training runs per environment were performed for the automated environment design. That is especially problematic considering that \ac{RL} in itself is considered to be computationally costly \cite{haarnojaSoftActorCriticOffPolicy2018a}. To reduce computation times, we want to discuss three options. First, our \ac{OPF} use case required multi-objective optimization. However, in most practical cases, we care about a single performance metric and can use single-objective optimization, which simplifies the problem. 
Second, while we considered overall \nDecisions different design variables, we found that only a few of them impact performance in significant ways. That suggests that the computation times can be reduced by using a smaller search space with only the most relevant design decisions.  
Third, since we utilized the \ac{HPO} framework, we can perform the optimization of the agent's hyperparameters together with the environment design variables. That can be expected to be more efficient than performing both subsequently. In summary, while our methodology is computationally expensive, there are multiple potential countermeasures. We leave them to future work. 
\par 

In section \ref{sec:related}, we noted that no general methodology for automated environment design exists so far. The work that fits best with ours is the observation space optimization by Kim and Ha \cite{kimObservationSpaceMatters2021}. While their algorithm was designed specifically for observations, our approach is usable for environment design on a general level and for multi-objective problems. To achieve that, we embedded our method in the existing \ac{HPO} framework to directly utilize existing algorithms, libraries, and best practices from the vast \ac{HPO} literature \cite{bischlHyperparameterOptimizationFoundations2023, eimerHyperparametersReinforcementLearning2023}.
Overall, to the best of our knowledge, our \ac{HPO}-based approach is the first general-level automated \ac{RL} environment design methodology of its kind. Together with Reda \etal \cite{redaLearningLocomoteUnderstanding2020}, it is also the most comprehensive analysis of environment design decisions for one specific domain.

\section{Conclusion}\label{sec:conclusion}

In this work, we proposed a general \ac{HPO}-based, automated, multi-objective \ac{RL} environment design methodology and applied it to the \ac{OPF} problem.
Using five open-source \ac{OPF} benchmark problems, we demonstrated how our automated design outperforms a comparable manually derived environment from literature.
We also showed how statistical analysis can be used to determine the design decisions that impact training performance in a statistically significant way. 
Our specific results 
uncovered multiple novel insights on how to formulate the \ac{OPF} problem as \ac{RL} environment.
Finally, we showed that resulting performance gains are reproducible for different experimental settings.
However, we also found a risk of overadjusting the environment to one specific \ac{RL} algorithm, similar to overfitting, which motivates further research on the topic.   
\par 
Our work has several implications for future research. 
First, the \ac{HPO}-based methodology is very general and can be applied to other \ac{RL} problems. Especially the robotics domain seems promising, considering that it is a main application of \ac{RL} with various options for environment design \cite{redaLearningLocomoteUnderstanding2020}.
Second, some high-performing design decisions found in this work may also apply to non-\ac{RL}-based \ac{OPF} approximation. For example, the \actionAutoscale and the mixture of training data may be directly transferable to supervised and unsupervised learning approaches \cite{parkCompactOptimizationLearning2024,liuNeuralNetworkApproach2024,huangUnsupervisedLearningSolving2024}. 
Finally, our methodology itself has some drawbacks that motivate future research. Especially its computational costs need to be reduced for better practical applicability.  

\begin{acks}
Many thanks to our colleague Rico Schrage and the anonymous reviewers for the valuable and constructive feedback. 
\end{acks}

\bibliographystyle{ACM-Reference-Format}
\bibliography{bibliography}

\appendix
\section{Hyperparameter Choices}

This section contains the exact parameters that were used for the \ac{RL} training runs in this work.

\begin{table}[H]
    \centering
    \small
    \caption{\ac{DDPG} and \ac{SAC} hyperparameter choices.}
    \label{tab:ddpgHPs}
    \begin{tabular}{lll}\toprule
         Hyperparameter & \ac{DDPG} & \ac{SAC} \\
         \midrule
         Actor neurons/layers & (256, 256, 256) & (256, 256, 256) \\
         Critic neurons/layers &  (256, 256, 256) & (256, 256, 256) \\
         Learning rate actor & 0.0001 & 0.0001 \\
         Learning rate critic & 0.0005 & 0.0005 \\
         Batch size & 256 & 256\\
         Gamma & 0.9 & 0.9 \\
         Memory size & 1000000 & 1000000 \\
         Noise standard deviation & 0.1 & N.A. \\
         Start train & 2000 & 2000  \\
         Tau & 0.001  & 0.001 \\
         Entropy learning rate & N.A. & 0.0001 \\
        \bottomrule
    \end{tabular}
\end{table}

\begin{table}[H]
    \centering
    \small
    \caption{The three specific environment designs used in this work.}
    \label{tab:baseDesign}
    \begin{tabular}{lllll}\toprule
        & Design Decision     & Base \cite{wolgastLearningOptimalPower2024} & Best Eco & Best Voltage   \\ \midrule
        Reward & \validReward      & 0.0 & 0.88 & 0.97 \\
        & \invalidPenalty           & 0.0 & 1.11 & 0.57\\
        & \textit{Invalid Obj. Share}  & 1.0 & 0.8 & 0.47\\ 
        & \penaltyWeight     & 0.1/0.5 & 0.54 & 0.16\\
        & \diffObjective       & False & True & True \\
        \midrule
        Data & \normalShare       & 0\% & 23.79\% & 35.51\% \\
        & \uniformShare      & 0\% & 41.24\% & 28.69\% \\ 
        & \simbenchShare     & 100\%  & 34.97\% & 35.8\% \\ 
        \midrule 
        Obs &  \textit{Add Voltage Mag.}           & False & False & False \\ 
        & \addVoltageAngle      & False & True & False \\
        & \addLineLoad        & False  & True & True \\
        & \addTrafoLoad    & False & True & True \\
        & \addSlackPower      & False & False & False \\
        \midrule
        Episode & \nSteps   & 1 & 1 & 1 \\ 
        \midrule
        Action & \actionAutoscale & True & True & True \\ 

        \bottomrule
    \end{tabular}
\end{table}

\section{Comprehensive Results}\label{app:results}

This section contains the full results for all five environments, including the dominated and non-dominated samples, the baseline environment design, and the statistically significant design decisions for all four evaluation criteria.

\subsection{Reactive Power Market}
\begin{figure}[H]
  \centering
  \includegraphics[width=1.0\linewidth]{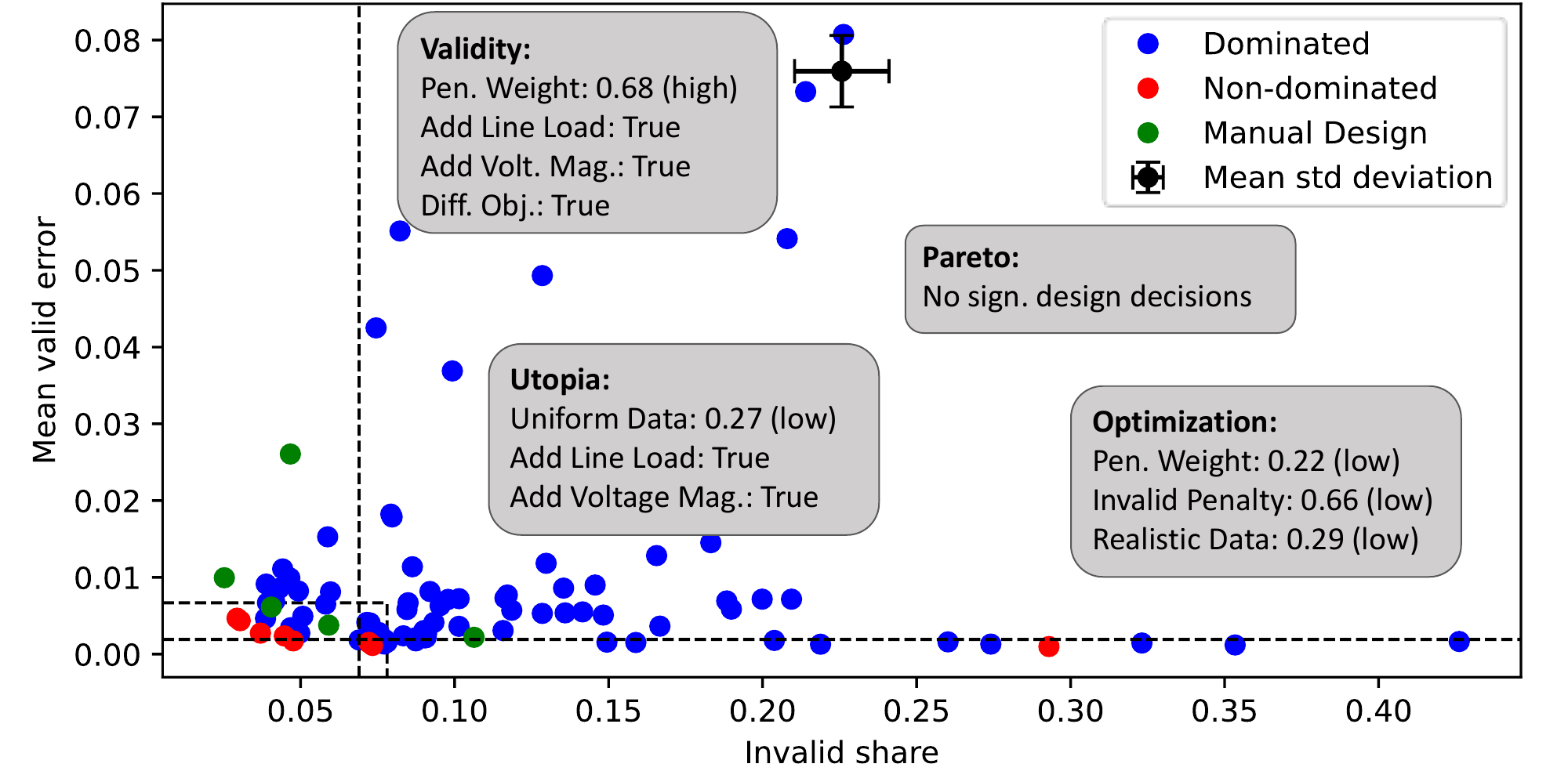}
  \caption{\qmarket Pareto-front and distribution of results, including all statistically significant environment design decisions.}
  \Description{TODO}
  \label{fig:qmarketAnnotated}
\end{figure}
\subsection{Maximize Renewable Feed-In}
\begin{figure}[H]
  \centering
  \includegraphics[width=1.0\linewidth]{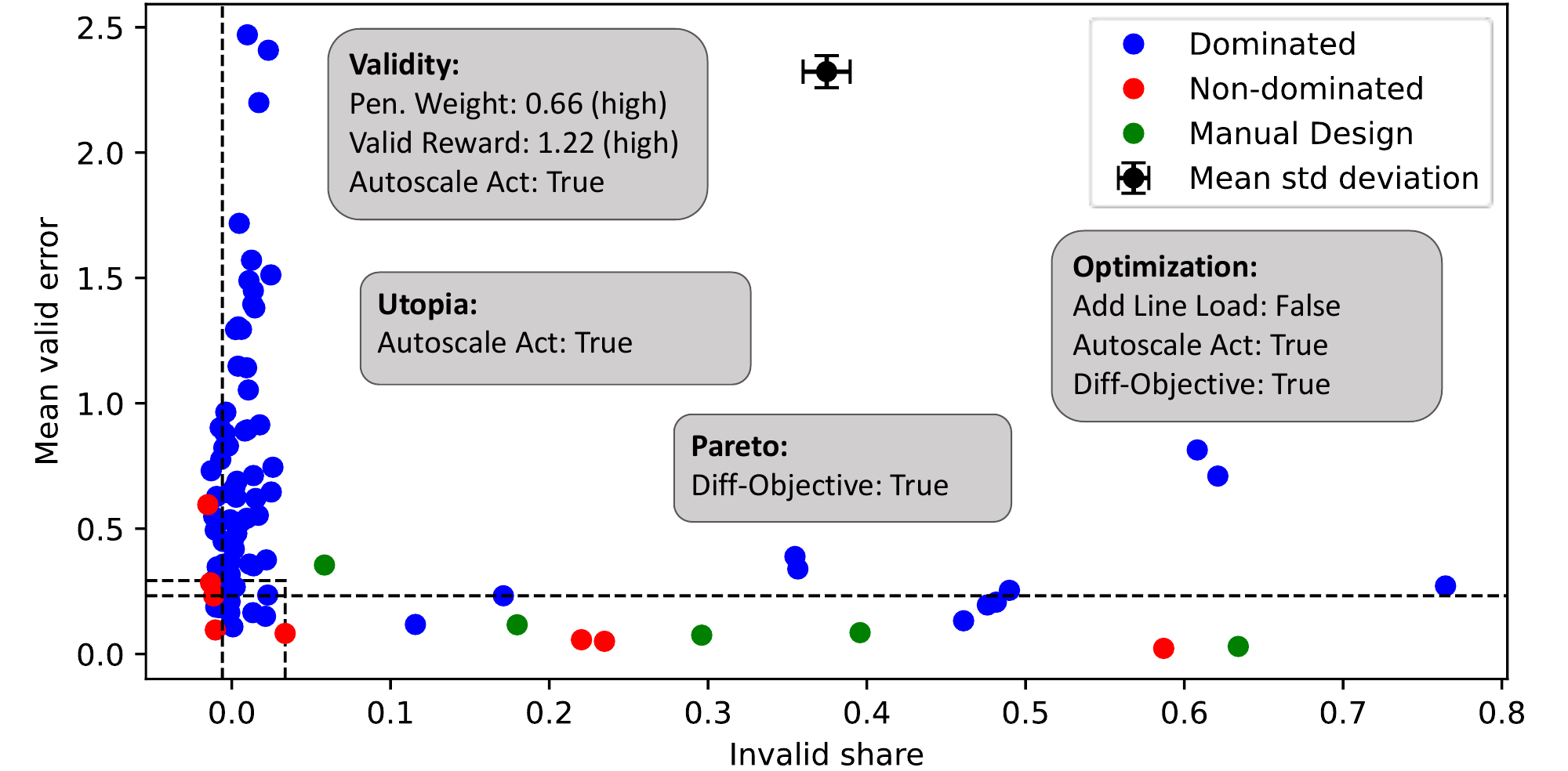}
  \caption{\maxRenewable Pareto-front and distribution of results, including all statistically significant environment design decisions.}
  \Description{TODO}
  \label{fig:renewableAnnotated}
\end{figure}
\subsection{Voltage Control}
\begin{figure}[H]
  \centering
  \includegraphics[width=1.0\linewidth]{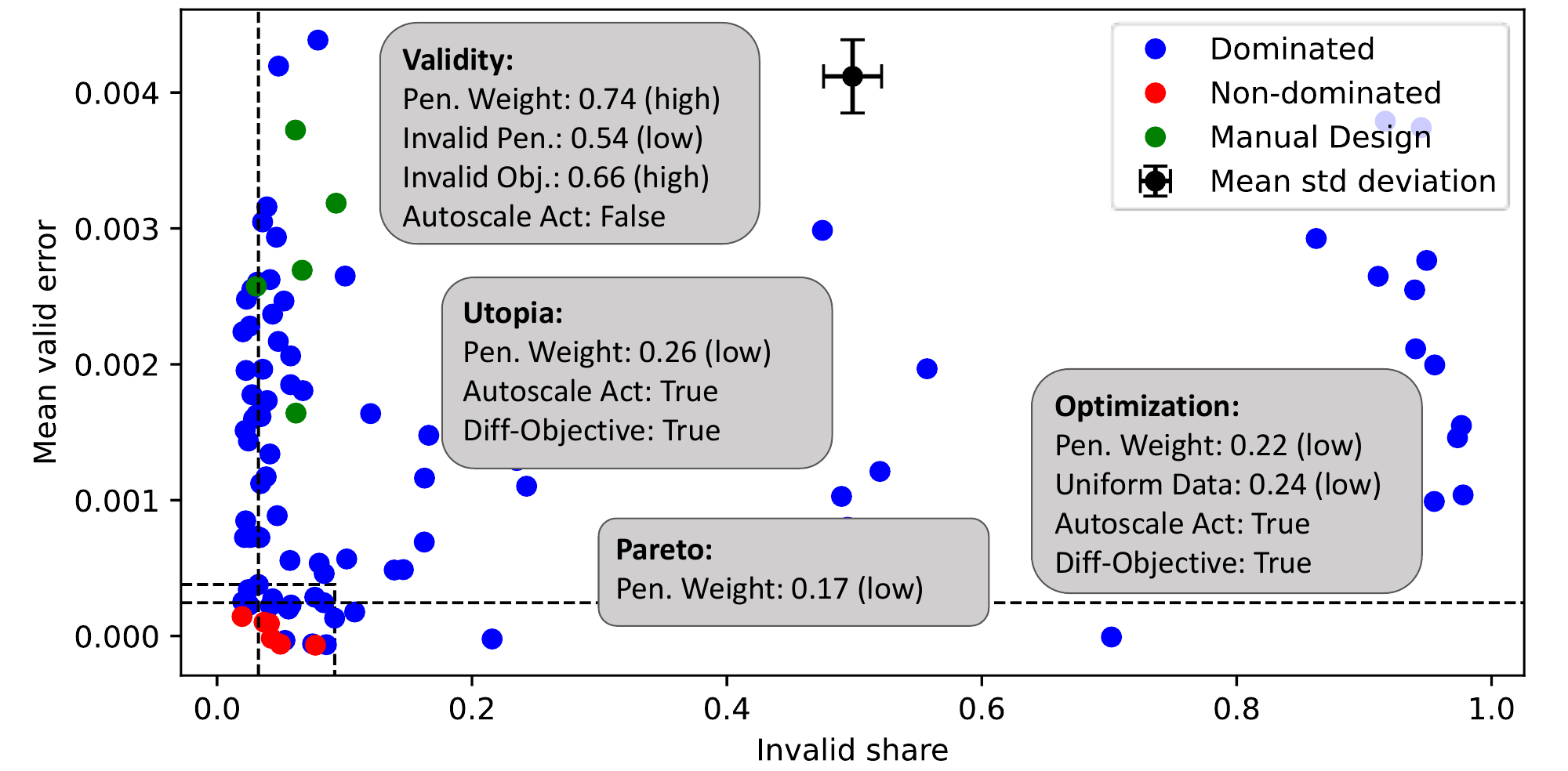}
  \caption{\voltageControl Pareto-front and distribution of results, including all statistically significant environment design decisions.}
  \Description{TODO}
  \label{fig:voltageAnnotated}
\end{figure}
\subsection{Economic Dispatch}
\begin{figure}[H]
  \centering
  \includegraphics[width=1.0\linewidth]{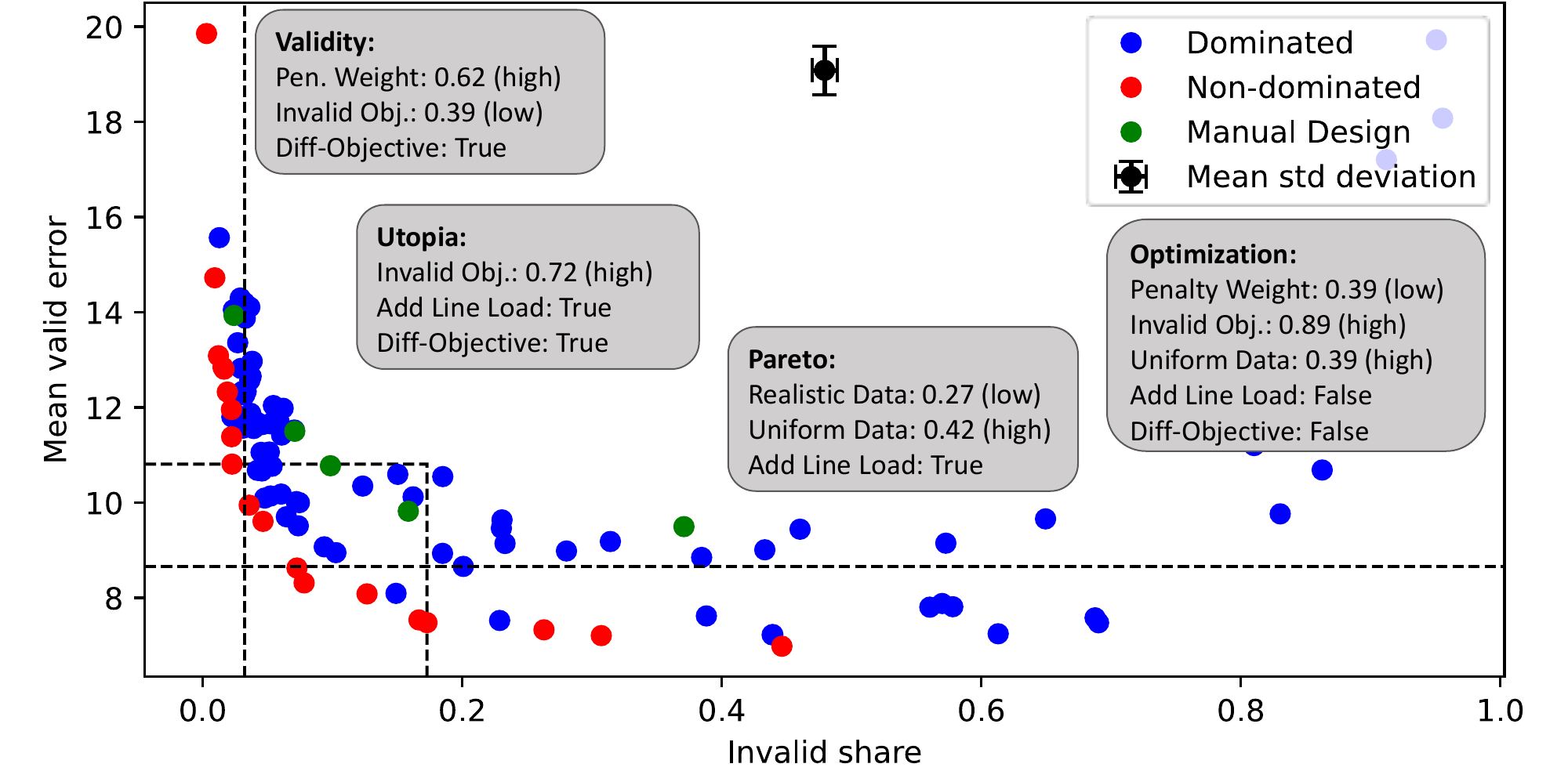}
  \caption{\ecoDispatch Pareto-front and distribution of results, including all statistically significant environment design decisions.}
  \Description{TODO}
  \label{fig:ecoAnnotated}
\end{figure}
\subsection{Load Shedding}
\begin{figure}[H]
  \centering
  \includegraphics[width=1.0\linewidth]{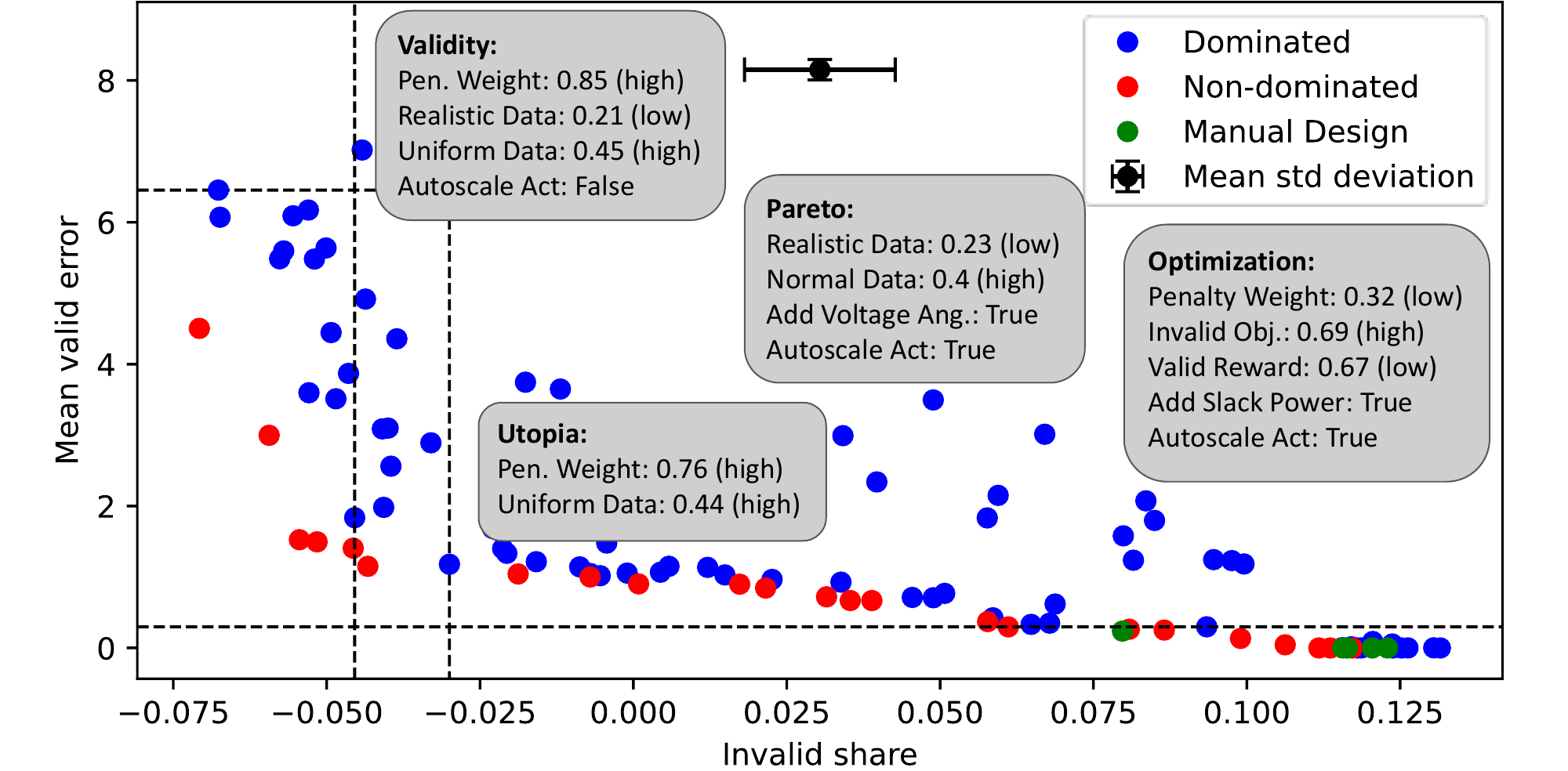}
  \caption{\loadShedding Pareto-front and distribution of results, including all statistically significant environment design decisions.}
  \Description{TODO}
  \label{fig:loadAnnotated}
\end{figure}

\end{document}